\documentclass{article}

    \PassOptionsToPackage{numbers, compress}{natbib}
    \usepackage[preprint]{neurips_2024}

\usepackage[utf8]{inputenc} % allow utf-8 input
\usepackage[T1]{fontenc}    % use 8-bit T1 fonts
\usepackage{hyperref}       % hyperlinks
\usepackage{url}            % simple URL typesetting
\usepackage{booktabs}       % professional-quality tables
\usepackage{amsfonts}       % blackboard math symbols
\usepackage{nicefrac}       % compact symbols for 1/2, etc.
\usepackage{microtype}      % microtypography

\hypersetup{
    colorlinks=true,
    linkcolor=blue,
    citecolor=blue,
    filecolor=cyan,
    urlcolor=purple
}

\usepackage[dvipsnames,svgnames,table]{xcolor}
\usepackage{graphicx}
\usepackage{amsmath}
\usepackage{amssymb}
\usepackage{booktabs}
\usepackage{multirow}
\usepackage{makecell}
\usepackage{caption}
\usepackage{subcaption}
\usepackage{bbm}
\usepackage{wrapfig}
\usepackage{xparse}
\usepackage{pifont}% http://ctan.org/pkg/pifont
\usepackage{algorithmic}
\usepackage[ruled,norelsize,linesnumbered]{algorithm2e}
\usepackage{enumitem}
\usepackage{pifont}% http://ctan.org/pkg/pifont
\usepackage{scalerel,xparse}
\usepackage{colortbl}

\usepackage{mathtools}
\usepackage{amsthm}
\usepackage{wasysym}

\NewDocumentCommand\emojione{}{\scalerel*{\includegraphics{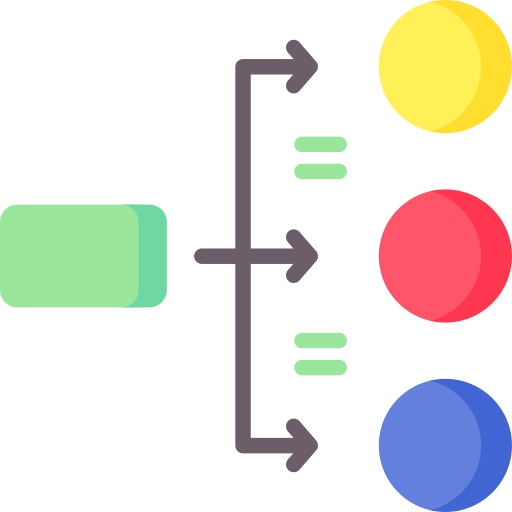}}{\rule{0.5cm}{0.7cm}}}

\title{
\vspace{-0.3cm}\emojione\,
Exploring and Improving Drafts in Blockwise Parallel Decoding
}

\author{%
  Taehyeon Kim$^{1,}$\thanks{Work done while at Google Research as a student researcher. Corresponding authors: <kimtaehyeon610@gmail.com> and <adbenton@google.com>.}\quad Ananda Theertha Suresh$^2$\quad Kishore Papineni$^2$\quad Michael Riley$^2$\\ \textbf{Sanjiv Kumar$^2$\quad Adrian Benton$^{2}$}
  \\
  $^1$KAIST AI \quad $^2$Google Research \\
}

\newcommand{\Autoref}[1]{%
  \begingroup%
  \def\algorithmautorefname{Algorithm}%
  \def\chapterautorefname{Chapter}%
  \def\sectionautorefname{Section}%
  \def\subsectionautorefname{Section}%
  \autoref{#1}%
  \endgroup%
}

\newcommand{\Tableautoref}[1]{%
  \begingroup%
  \def\figureautorefname{Table}%
  \autoref{#1}%
  \endgroup%
}

\newcommand{\removed}[1]{}
\newcommand{\REMOVED}[1]{}

\newcommand{\ngram}{\textit{n}-gram\ }
\newcommand{\ngrams}{\textit{n}-grams\ }

\begin{document}

\maketitle
\begin{abstract}
\vspace{-5pt}
Despite the remarkable strides made by autoregressive language models, their potential is often hampered by the slow inference speeds inherent in sequential token generation.
Blockwise parallel decoding (BPD) was proposed by Stern et al.\,\cite{bpd} as a method to improve inference speed of language models by simultaneously predicting multiple future tokens, termed \textit{block drafts}, which are subsequently verified and conditionally accepted by the autoregressive model.
%selectively accepted by the autoregressive model.
%Block drafts are generated by multiple independent prediction heads of blockwise parallel language models.
This paper contributes to the understanding and improvement of block drafts in two ways. First, we analyze the token distributions produced by multiple prediction heads. Secondly, we leverage this analysis to develop algorithms to improve BPD inference speed by refining the block drafts using \ngram and neural language models.
Experiments demonstrate that refined block drafts yield a +5-21\% increase in block efficiency (i.e., the number of accepted tokens from the block draft) across diverse datasets. 
% higher average verified prefix length across diverse datasets.
%, achieving a 5--21\% increase in block efficiency across diverse datasets. 
\end{abstract}

\vspace{-10pt}
\section{Introduction}\label{sec:intro}
%\vspace{-6pt}
% Emergence of Large Language Models and Limitations on the Applications
The landscape of natural language processing has been profoundly reshaped by recent advances in autoregressive language models \cite{gpt3, emergent_abilities_llms, lms_are_unsupervised_multitask_learners, t5, gemini}.  These models have shown remarkable proficiency across a range of text generation tasks, including applications like question answering \cite{squad} and summarization \cite{cnn_sum}. However, a significant obstacle to their wider application is high inference latency, particularly for extremely deep models with hundreds of billions of parameters \cite{chinchilla, gopher, palm}. This latency, intrinsic to decoding with autoregressive language models (LMs), imposes considerable computational burdens and limits real-time deployment.

% Introduce related works: Speculative Decoding and Blockwise Parallel Decoding
In response to these latency challenges, the field has seen a shift towards decoding methods aimed at reducing the inference latency in large language models (LLM). One promising development is the concept of blockwise parallel decoding (BPD) \cite{bpd, pass, medusa}. Unlike autoregressive decoding, which generates one token at a time, blockwise parallel LMs are outfitted with a set of prediction heads that propose and verify a draft, a block of subsequent tokens, in parallel. While BPD offers one solution to accelerated text generation, it also poses a challenge in ensuring that the proposed drafts are fluent and natural.

\begin{figure*}
    \centering
    \begin{subfigure}{0.4\textwidth}
        \includegraphics[width=\linewidth]{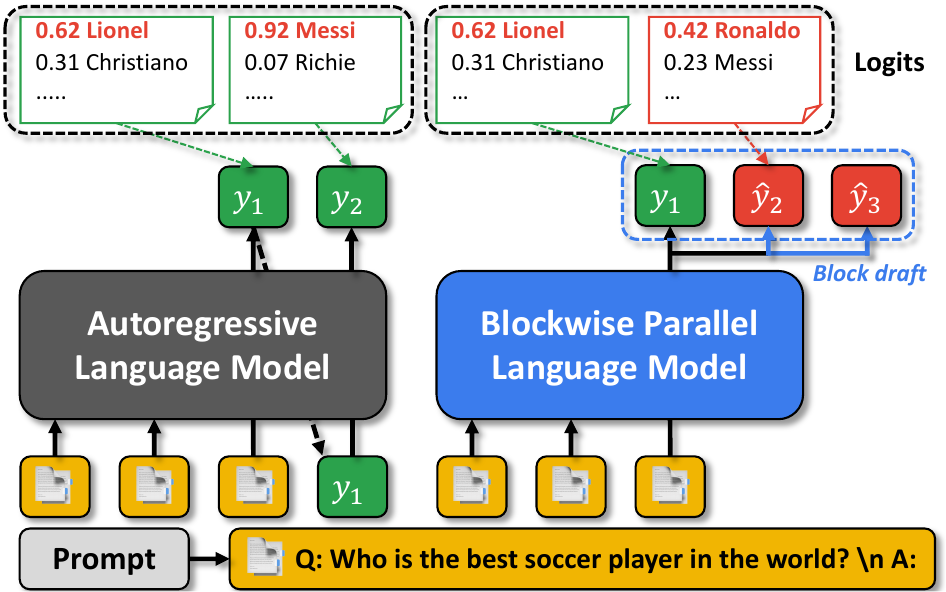}
        \caption{Example of block drafts}     \label{fig:example}
    \end{subfigure}
    \begin{subfigure}{0.59\textwidth}
        \includegraphics[width=\linewidth]{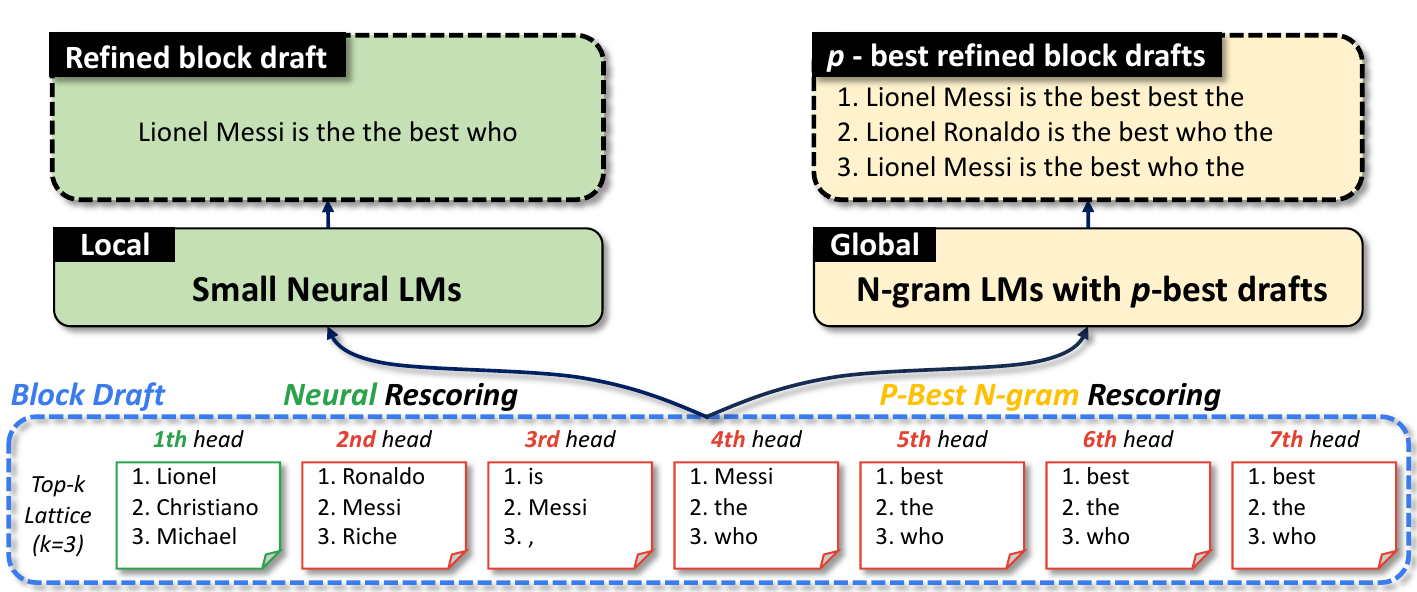}
        \caption{Output of our proposed algorithms} \label{fig:contri}
    \end{subfigure}
    \caption{
    % \adrian{I would remove mention of "Finite State Transducer" from the diagram. We never actually use any transducers in our experiments. Maybe "N-gram LMs" (with p-best rescoring?)} 
    (a) Illustration of two tokens that are decoded by autoregressive decoding vs. two tokens drafted by BPD. (b) Outputs from our proposed algorithms, where the top-$k$ token-level predictions are refined using local neural and global \ngram rescoring, which selects the $p$ most probable sequences by dynamic programming, for batched verification. 
    % \theertha{In Figure (a), I am not sure what is autoregressive models doing there? Would it be better to have 3 figures e.g., (a) autoregressive (b) block draft and (c) proposed algorithm? }\taehyeon{@Taehyeon: Figure 1 (a) is a just colorized version of the previous figure (submitted to ICML). We don't have enough spaces to 3 figures in a row... btw, do you think the autoregressive figure does not captures the nature of autoregressive decoding?}
    }
    \vspace{-5pt}
\end{figure*}

% Explain the things more details with an example (Figure) such as Who is the best soccer player? A: 
BPD inference speed depends on both the time it takes to produce a block draft and the draft's agreement with the base LM's output (\autoref{fig:example}). Unlike standard autoregressive LMs that generate tokens sequentially—ensuring consistency with all preceding tokens (e.g., `Messi' following `Lionel')—BPD employs a parallel strategy. Here, blockwise parallel LMs simultaneously predict multiple token drafts (e.g., `Lionel' and `Ronaldo'), each independently. The primary challenge in BPD is ensuring that these concurrently generated tokens maintain consistency. Effective block drafters should prioritize coherent sequences such as `Lionel Messi' over less likely combinations like `Lionel Ronaldo', which a robust LM would not decode. The focus of this research is improving the quality of block drafts without altering the underlying model parameters.
% BPD inference speed is a function of the time required to produce a block draft, and how well that draft matches the decode produced by the base LM (\autoref{fig:example}). While standard autoregressive LMs generate tokens sequentially, ensuring that each generated token is consistent with all of the preceding tokens (e.g., `Messi' is consistent with `Lionel'), BPD adopts a different strategy. At each time step, blockwise parallel LMs predict a draft of multiple tokens concurrently (`Lionel' and `Ronaldo'), each of which is predicted independently of the others. The inherent challenge in block drafts lies in maintaining consistency between tokens at each position during parallel generation. A good block drafter should prioritize congruous sequences, like `Lionel Messi', over incongruous sequences, like `Lionel Ronaldo', which are unlikely to be decoded by a strong LM. This work focuses on improving the quality of block drafts without modifying the original model parameters.

\vspace{-10pt}
\section{Our contributions}\label{sec:contri}
\vspace{-5pt}
In this paper, we first investigate properties made by the prediction heads of blockwise parallel LMs across several tasks; given these observations, we propose rescoring algorithms to produce higher quality block drafts.

\vspace{-5pt}
\subsection{Observations on block drafts}
\vspace{-2pt}

\textbf{Consecutive repetitions} \,All heads within a block make predictions independently in a blockwise parallel LM. Unsurprisingly, we observe that this leads to block drafts with significant token repetition across heads. Consecutive repetition is pervasive across tasks, ranging from 20\% to 75\% of all neighboring draft tokens, depending on the task (\textbf{\Autoref{subsec:consec_repetition}}).

\textbf{Confidence of different heads} \,We analyze the distribution of probabilities within each softmax head. Our empirical analysis reveals a key property of BPD: the block drafter tends to be more confident with initial tokens, and becomes progressively less confident for subsequent tokens. We find that the confidence of block heads correlates strongly with the quality of the block drafter (\textbf{\Autoref{sec:cascade}}).

\textbf{Oracle top-$k$ block efficiency} \,In the standard BPD algorithm (\textbf{\hyperref[alg:bpd]{Algorithm 1}}), the most likely token at each head is generated as the draft. As highlighted before, this approach is prone to two issues: (1) there might be consecutive repetitions and (2) the model might not be confident about the prediction at some of the heads. 
We use block efficiency, the average number of draft tokens accepted during decoding, to measure the quality of a given drafter \cite{speculative_decode, spectr}.
We ask if the block efficiency can be improved by considering the top-$k$ most likely tokens at each head. To measure the potential benefit of considering top-$k$ tokens, we measure the block efficiency of the oracle path through this top-$k$ lattice, \emph{oracle top-$k$ block efficiency}, and show that there is significant headroom for improvement across tasks (\textbf{\Autoref{sec:oracle}}).

\vspace{-5pt}\subsection{New algorithms}
\vspace{-2pt}
Based on these observations, we propose two algorithms to leverage the top-$k$ predictions at each head and improve BPD latency\,(\Autoref{fig:contri}). Neither of these algorithms require changes to the underlying blockwise parallel LMs.
% In order to capitalize on top-$k$ predictions at each head, we propose two algorithms. 

\textbf{Local rescoring via neural LMs}  \,Given the top-$k$ predictions at each head, we refine the block draft by using a small neural, autoregressive LM to greedily rescore these local predictions (\textbf{\Autoref{sec:neural}}).
While the block prediction scores are produced independent of each other, neural rescoring should favor sequences that are fluent, encouraging coherence between the predictions at each head.

\textbf{Global rescoring via \ngram LMs with multi-drafts} \,
If the blockwise parallel LM has $h$ heads and we consider the top-$k$ tokens from each head, then there are $k^h$ candidate drafts of length $h$ that can be formed. We propose to use an \ngram model to efficiently rescore \textit{all} paths, via dynamic programming, and generate the $p$ most probable rescored paths as a batch of draft candidates. These $p$ drafts can then be verified in parallel by the blockwise parallel LM (\textbf{\Autoref{sec:ngram}}).

\begin{wrapfigure}{r}{0.45\textwidth}
    % \vspace{-15pt}
    \includegraphics[width=1.0\linewidth]{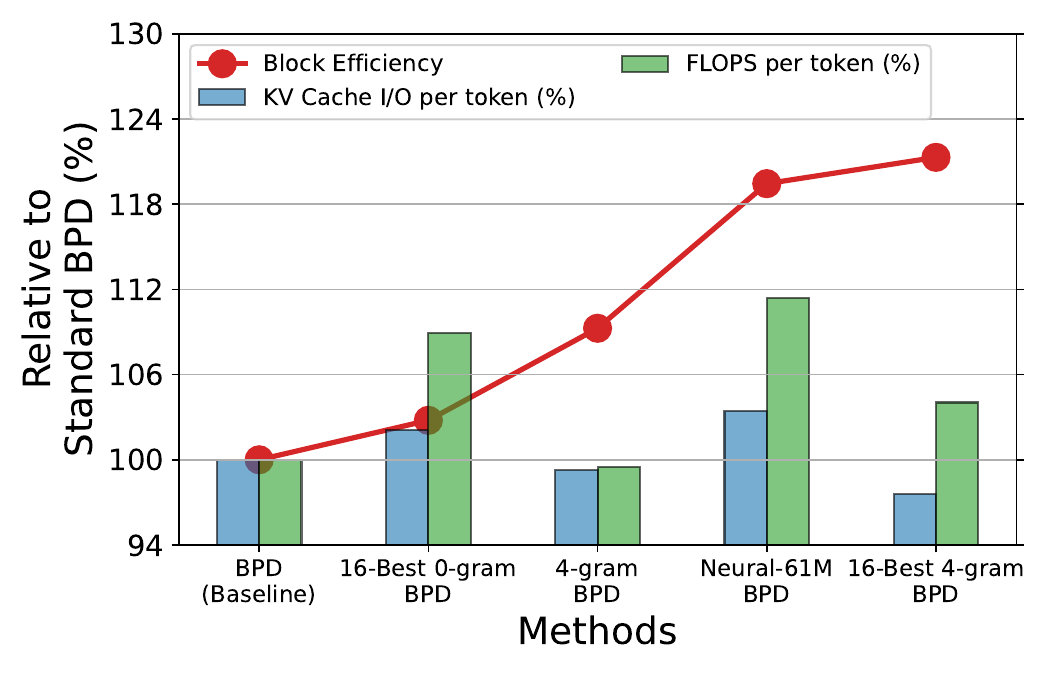}
    \vspace{-18pt}
    \caption{
    % \adrian{Right-hand y-axis should be something like "Cache I/O \& FLOPS relative to Autoregressive (\%)". None of those values are increases over baseline. I would also drop "FST" and replace it with "16-best", or similar.} 
    Relative performance of our methods to standard BPD with a 1.5B parameter blockwise parallel LM on NewsRoom dataset \cite{newsroom}. 
    % Block efficiency gains are given absolute. \% change in KV Cache I/O and FLOPS per token are relative to standard autoregressive decoding to our BPD variants. 
    Details are described in \textbf{\Autoref{sec:app_memory}}.} 
    \vspace{-15pt}
    \label{fig:overall_intro}
\end{wrapfigure}

There are two critical distinctions between the proposed algorithms: the amount of context/expressive power available to each class of rescoring model, and fundamental limitations of decoding with each class. While neural rescoring models are potentially more expressive and can leverage unbounded context, \ngram{} LMs can be used to efficiently find the globally most likely rescored drafts from the exponentially-sized set of possible draft candidates. 
{\Autoref{fig:overall_intro} shows that our proposed methods enhance block efficiency, with one approach increasing it by up to \textbf{+21.30\%}. This same method also optimizes resource usage, reducing key-value (KV) cache I/O by \textbf{-2.54\%} and additionally using FLOPs per token by \textbf{+4.04\%}.}
% \Autoref{fig:overall_intro} shows that our methods boost the block efficiency (up to \textbf{+21.30\%}) with little additinal use of key-value (KV) cache I/O (-2.54\%) and FLOPs per token (+4.04\%).
Description of each algorithm is given in \textbf{\Autoref{sec:algorithms}}.

\vspace{-5pt}
\subsection{Organization}
\vspace{-5pt}
The remainder of this paper organized as follows. In \textbf{\Autoref{sec:related}}, we discuss previous literature in reducing LLM latency. In \textbf{\Autoref{sec:prelim}}, we define foundational concepts and terminology. \textbf{\Autoref{sec:exp_set}} describes our experimental setup, datasets, on methods. \textbf{\Autoref{sec:explore}} describes our analysis of the block drafts. In \textbf{\Autoref{sec:algorithms}}, we present the proposed BPD rescoring algorithms and empirical results, followed by a final discussion in \textbf{\Autoref{sec:conclu}}.

\vspace{-10pt}
\section{Related Work}\label{sec:related}
\vspace{-5pt}
\subsection{Efficient transformer inference}
\vspace{-5pt}
Works on improving transformer efficiency encompass both optimization of an existing set of model weights, or a fundamental change to the model architecture. Examples of the former include techniques such as quantization \cite{xiao2023smoothquant, yao2022zeroquant, dettmers2022gpt3} and model pruning \cite{sun2023simple, ma2023llm}. In parallel, neural architecture search has played a crucial role in identifying network structures that balance performance with efficiency \cite{kitaev2020reformer, zhu2021long}. Relatedly, Elbayad et al.\,\cite{elbayad2019depth} propose early-exiting at intermediate layers for faster inference, while Schuster et al.\,\cite{calm} explore confidence thresholding for balancing speed and accuracy. These methods offer insights into optimizing decoding under resource constraints.

One important line of work has focused on modifying the decoding method in LMs. The adoption of non-autoregressive (parallel) decoding strategies \cite{bpd, nonar} marks a pivotal shift in this domain, addressing inference latency by simultaneously generating multiple tokens. Subsequent innovations have sought to refine this approach by incorporating additional context \cite{chi2020align}, iterative refinement \cite{biglittlespec}, and tree-based attention mechanism \cite{medusa}. However, these refinements often require complex training or additional inference data.

\vspace{-5pt}
\subsection{Efficient and effective decoding}
\vspace{-5pt}

There are several recent works that improve the speed of LLM decoding, including pioneering works like BPD and speculative decoding. Speculative decoding leverages a smaller `draft' model to anticipate the outputs of a larger target model, improving average decode latency without loss in generation quality \cite{speculative_decode, speculative_decode2, biglittlespec}. The draft model is typically trained on the same corpus as the LLM, thus autoregressively generates similar drafts as the target model with reduced latency. Speculative decoding is most successful when a long sequence of speculated tokens are accepted by the target LM during verification, avoiding multiple serial calls to the target LM to generate the same sequence.

On the surface, contrastive decoding algorithms share some similarities with our proposed draft rescoring approach, insofar as a weaker model is used to modify the predictions of the target LM \cite{contrastive_decoding, instructive_decoding}. However, in this work, we refine block drafts solely to improve latency. Like speculative decoding, our proposals have no effect on the quality of the target LM's generated text.

\vspace{-10pt}
\section{Preliminaries}\label{sec:prelim}
\vspace{-5pt}
This section introduces notation and concepts, including algorithms for standard autoregressive decoding and BPD.

\textbf{Autoregressive decoding}\, Let \( \mathcal{M}_{\theta} \) be an autoregressive LM parameterized by \( \theta \). The objective is to generate an output sequence \( y_{\leq T} = (y_1, \ldots, y_T) \) conditioned on an input sequence $\bar{x}$.
 $z_t = \mathcal{M}_{\theta}(y_{t} | \bar{x}, y_{< t})$ is a vector of logits, $z_t \in \mathbb{R}^{|\mathcal{V}|}$, where $\mathcal{V}$ is the vocabulary over tokens. These logits define a conditional probability distribution at each time step $p_\theta(y_{t+1} | \bar{x}, y_{\leq t}) = \frac{e^{z_t}}{\sum e^{z_t}}$, which by the chain rule yields $p_{\theta}(y_{\leq T}| \bar{x}) = \prod_{t=1}^{T} p_{\theta}(y_{t}| \bar{x}, y_{< t})$.

Sequences are generated autoregressively, either through ancestral sampling from some form of the conditional next token distribution \cite{nucleus_sampling}, or by a beam search through the space of possible sequences to return a probable sequence. Greedy decoding, a special case of beam search, generates each token as $\hat{y}_{t+1} = \operatorname*{arg\,max} p_{\theta}(y_{t+1}| \bar{x}, y_{\leq t})$. In this work, we consider greedy decoding exclusively, as this is the setting that Stern et al.\,\cite{bpd} was designed to accelerate.

\begin{figure}[t]
\begin{algorithm}[H] 
\DontPrintSemicolon
    \caption{Blockwise parallel decoding (BPD)}
    \label{alg:bpd}
    \begin{algorithmic}[1]
    \SetKwInOut{Input}{Input}
    \INPUT: Blockwise parallel LM $\mathcal{M}_{\theta}^h$, initial prompt sequence $\bar{x}$ and target sequence length $T$.\\
    \STATE Initialize $t \leftarrow 1$ \\
    \WHILE{$t < T$} 
        \STATE \textcolor{blue}{\textbf{/* Stage 1: Predict */}}
        \STATE $z^1_{t,1}, \dots, z^h_{t,h} \leftarrow   \mathcal{M}_{\theta}^h(y_{t+1}, \dots, y_{t+h} |\bar{x}, y_{\leq t})$
        \STATE $y_{t+1}, \hat{y}_{t+2}, \dots, \hat{y}_{t+h} \leftarrow  \operatorname*{arg\,max}(z^1_{t,1}), \operatorname*{arg\,max}(z^2_{t,2}), \cdots, \operatorname*{arg\,max}(z^h_{t,h})$
        % \STATE $\hat{y}_{t+2}, \dots, \hat{y}_{t+h} \leftarrow $ $\operatorname*{arg\,max}(z^2_{t,2}), \cdots, \operatorname*{arg\,max}(z^h_{t,h})$
        \STATE \textcolor{blue}{\textbf{/* Stage 2: Verify */}} \\
        \FOR{$j \leftarrow 2, \dots, h$ in parallel}
            \STATE $z_{t,j}^1, \cdots \leftarrow $$\mathcal{M}_{\theta}^h(y_{t+j}, \dots |\bar{x}, y_{\leq {t+1}}, \hat{y}_{t+2}, \cdots, \hat{y}_{t+j-1})$
        \ENDFOR
        \STATE \textcolor{blue}{\textbf{/* Stage 3: Accept */}} \\
        \STATE $n \leftarrow \max \{n: \hat{y}_{t+j} = \operatorname*{arg\,max} z^1_{t, j}, 2 \leq j \leq n \}$ \\
        \STATE $t \leftarrow t + n  $ \\
    \ENDWHILE
\end{algorithmic}
\end{algorithm}
\vspace{-10pt}
\end{figure}

\textbf{Blockwise parallel decoding}\, Let \( \mathcal{M}^h_{\theta} \) be a blockwise parallel LM with block size $h$. This model employs \( h \) distinct feedforward neural (FFN) layer with a single hidden layer, atop the target LM's final hidden layer. The output of each FFN is followed by a softmax layer over the vocabulary to predict each of the \( h \) subsequent tokens in the block. In our experiments, the parameters of the FFNs are learned jointly with the base LM during training, and the weights of all softmax layers are tied to the input embedding table. \textbf{\hyperref[alg:bpd]{Algorithm 1}} describes the BPD greedy decoding procedure:

\begin{enumerate}
\vspace{-5pt}
    \item \textbf{Predict:} \( \mathcal{M}_{\theta}^h \) is used to generate a draft of \( h \) token predictions \( y_{t+1}, \hat{y}_{t+2} \cdots, \hat{y}_{t+h} \), conditioned on the prompt, \( \bar{x} \), and existing generated text, \( y_{\leq t} \). $y_{t+1}$ is identical to the target LM greedy decode.
    %These tokens are generated conditioned on the input prompt, \( \bar{x} \), and the existing generated text, . 
    % \theertha{softmax or classification heads?}
    
    \item \textbf{Verify:} At this stage, the target LM greedily generates next-token logits \( \{ z_{t,2}^1, \cdots, z_{t, h}^1 \} \) conditioned on the existing prefix and block draft \( \{ \bar{x}, y_{\leq t+1}, \hat{y}_{t+2}, \cdots, \hat{y}_{t+h} \} \). Verification amounts to checking which block draft tokens match the autoregressive greedy decode from the target LM: ($\operatorname*{arg\,max}{z^1_{t,i}}) == \hat{y}_{t+i}$. Verification of all positions can be performed in parallel under the assumption that the target LM is a decoder-only transformer.
    %This verification ensures that only a subset of the predicted tokens aligns with predictions of the base model (i.e., 1st head), guaranteeing that the final sequence remains consistent with autoregressive decoding. 
    % Only the matching tokens are accepted and integrated into the final sequence in the subsequent stage.
    % The verification checks whether the predicted token is identical to the tokens from the base model (i.e., 1st head). This critical step ensures that only contextually appropriate predictions are retained for the final sequence, thereby maintaining the performance of the base model (i.e., autoregressive decoding with 1st head) within the BPD framework.
    
    \item \textbf{Accept:} Finally, the length of the longest contiguous prefix of draft tokens that match the target LM greedy decode is identified: $n$. The decoded sequence is extended by $n+1$ tokens and we iterate.\footnote{The decoded sequence is extended by $n+1$ tokens since during verification we generate the token from the target LM, $\operatorname*{arg\,max}{z^1_{t,n+1}}$, at the first position where the draft differs from the target LM greedy decode.} Note that in general, not all $h$ tokens are accepted, and many of the draft tokens in each block are discarded. Since the additional time required to generate a block of tokens is fast relative to the time it takes for the forward pass of the target LM, a modest gain in accepted prefix length justifies the cost of draft generation.
    % This stage ensures the model's efficiency without compromising on the quality of the generated text.
\end{enumerate}

% \theertha{I think we need to highlight that in general not all $k$ tokens are accepted.}

% This paper delves into the intricacies of refining BPD, aiming to harmonize swift token generation with linguistic coherence. Our empirical investigations reveal that BPD naturally orchestrates the confidence of its predictions in a sequentially diminishing fashion, a phenomenon we term \textit{hierarchical confidence ordering}. 
% % \theertha{In my mind, hierarchical means there is a tree type of structure, one very confident and few intermediate nodes which are not so confident and then few leaves which are very less confident. As far as I can see there is no tree structure here. Would it make sense to call this phenomenon something else? e.g., sequential confidence ordering? (I don't like this name too... } 
% This observed pattern serves as a crucial insight into the model's predictive behavior, allowing for strategic adjustments that can potentially enhance oracle block efficiency, particularly in tasks requiring sequential predictions. \theertha{Both Oracle and block efficiency are not defined yet, perhaps we need to introduce it before stating here? }  \theertha{I am not fully sure I follow how we use these insights to design our algorithms. Can you please add a few lines about it here?}

\vspace{-15pt}
\section{Experimental setup} \label{sec:exp_set}
\vspace{-5pt}
\begin{figure}[ht]
\centering
\begin{minipage}{0.43\textwidth}
\centering
\vspace{-1.5pt}
\captionof{table}{Per-task test performance of each finetuned model and block efficiency over language modeling (LM), extractive question answering (QA), and both long and short summarization (L-Sum \& S-Sum).  }\label{tab:baseline}
%Tasks include Language Modeling (LM), extractive Question Answering (QA), and both Long and Short Summarization (L-Sum \& S-Sum). 
% The performance metric for LM is perplexity, for QA is exact match, and for the remaining summarization tasks, the metric is ROUGE-L.

\vspace{-3pt}
\resizebox{\textwidth}{!}{%
\begingroup
\setlength{\tabcolsep}{2.8pt} % Default value: 6pt
\renewcommand{\arraystretch}{0.95}
\tiny
\begin{tabular}{@{}c|l|c|c@{}} 
\toprule
 \multirow{2}{*}{Task} & \multirow{2}{*}{Dataset} & \multirow{2}{*}{Performance} & Block \\ 
 & & & Efficiency \\\midrule
LM                   & LAMBADA \cite{lambada}  & 7.88 & 3.12 \\ \midrule
QA                   & SQuAD V1 \cite{squad}       & 57.60  & 2.08 \\ \midrule
\multirow{2}{*}{S-SUM} & CNN/Daily \cite{cnn_sum}     & 39.85  & 1.74 \\
                     & SAMSUM \cite{SAMSum}        & 37.66  & 1.27 \\ \midrule
\multirow{3}{*}{L-SUM} & MultiNews \cite{multi_news}     & 23.08  & 1.10 \\
                     & XSUM \cite{xsum}          & 52.15  & 1.13 \\ 
                     & NewsRoom \cite{newsroom}      & 39.85  & 1.08 \\\bottomrule
\end{tabular}
\endgroup
}
\end{minipage}
\hspace{0.3cm}
\begin{minipage}{0.53\textwidth}
\centering
\vspace{-1.5pt}
\captionof{table}{Outputs from BPD frameworks. % on LAMBADA, SQuAD V1, and CNN/Daily datasets. 
Black indicates standard decoded output, \textcolor{blue}{blue} indicates accepted draft tokens, and \textcolor{brown}{brown} is the prompt.}
\vspace{-3pt}
\label{tab:case}
\scriptsize
\begingroup
\setlength{\tabcolsep}{2.8pt} % Default value: 6pt
\renewcommand{\arraystretch}{1.0}
\begin{tabular}{@{}p{\linewidth}@{}}
\toprule
\textbf{LAMBADA} \\ \midrule
\textcolor{blue}{it's} no\textcolor{blue}{thing more than} a faceless, formless \textcolor{blue}{brown blob} to me, \textcolor{blue}{but I} take \textcolor{blue}{his word} \textcolor{blue}{for the} resemblance to our \textcolor{blue}{genetic makeup.} \textit{\textcolor{gray}{... \{Skip\}...}} \\\midrule
\textbf{SQuAD V1} \\ \midrule
\textcolor{brown}{Question: Who was announced as the LEM contractor in November 1962? context: Wiesner kept up the pressure, even making the disagreement public} \textit{\textcolor{gray}{... \{Skip\}...}} \\
\textcolor{brown}{Answer:} Gru\textcolor{blue}{mman}
\\\midrule
\textbf{XSUM} \\\midrule
\textcolor{brown}{Summarize:} \textit{\textcolor{gray}{... \{Skip\}...}} \\
\textcolor{blue}{Millions of} small businesses will benefit from a \textcolor{blue}{reduction of} business rate from the Budget Osborne, Chancellor George \textcolor{blue}{Osborne has announced.}
\\ \bottomrule
\end{tabular}
\endgroup
\end{minipage}
\vspace{-5pt}
\end{figure}

% \begin{table}[t] \scriptsize \addtolength{\tabcolsep}{1.2pt}
% \centering
% \caption{Per-task test performance of each finetuned model and block efficiency. Tasks include Language Modeling (LM), extractive Question Answering (QA), and both Long and Short Summarization (L-Sum \& S-Sum). The performance metric for LM is perplexity, for QA is exact match, and for the remaining summarization tasks, the metric is ROUGE-L.}
% \label{tab:baseline}
% \begin{tabular}{@{}c|l|c|c@{}} 
% \toprule
%  \multirow{2}{*}{Task} & \multirow{2}{*}{Dataset} & \multirow{2}{*}{Performance} & Block \\ 
%  & & & Efficiency \\\midrule
% LM                   & LAMBADA \cite{lambada}  & 7.88 & 3.12 \\ \midrule
% QA                   & SQuAD V1 \cite{squad}       & 57.60  & 2.08 \\ \midrule
% \multirow{2}{*}{S-SUM} & CNN/Daily \cite{cnn_sum}     & 39.85  & 1.74 \\
%                      & SAMSUM \cite{SAMSum}        & 37.66  & 1.27 \\ \midrule
% \multirow{3}{*}{L-SUM} & MultiNews \cite{multi_news}     & 23.08  & 1.10 \\
%                      & XSUM \cite{xsum}          & 52.15  & 1.13 \\ 
%                      & NewsRoom \cite{newsroom}      & 39.85  & 1.08 \\\bottomrule
% \end{tabular}
% \end{table}

In this paper, we use $\approx1.5$ billion (B) parameter decoder-only transformer LMs with up to 9 blockwise heads.\footnote{This study is based on the original BPD framework, with a modification: we use decoder-only models instead of the T5 encoder-decoder architecture. Other setups are consistent with the approach in Stern et al.\,\cite{bpd}.}
The 1.5B model and all other LMs were pretrained on (English) C4 \cite{c4} with the causal next token prediction objective tokenized with the GPT3 subword vocabulary \cite{gpt3}. For the 1.5B blockwise parallel LMs, all heads were trained jointly to predict the following $h$ tokens at each iteration. During pretraining, we use batches of 2048 subword sequences, each 512 tokens in length, amounting to $\approx 200$B input tokens in total on TPUv3/TPUv4 \cite{jouppi2017datacenter} with Jax \cite{jax2018github}.
% \adrian{Should we mention using TPU/Jax? Am afraid this will be a tell for affiliation. Given that we do not report wall clock times, I would argue that the hardware we use, and the library we used to define the neural network was irrelevant.}

We evaluate the potential latency improvement of block drafts by \textit{block efficiency} \cite{speculative_decode, spectr}. In this context, block efficiency represents the theoretical speedup compared to standard greedy decoding. It is defined as the average number of tokens decoded per serial call to the blockwise parallel LMs. The formula for block efficiency is given by $ B \coloneqq \frac{\text{Total number of decoded tokens}}{\text{Total number of serial calls to } \mathcal{M}_{\theta}^h}$.

In this definition, the total number of decoded tokens is the sum of the number of accepted tokens across decoding steps, not necessarily all $h$ predicted tokens in each block. Only the tokens that pass the `Verify' stage and align with the base model's predictions are accepted and integrated into the final sequence. This ensures that generated text is identical to the target LM, while achieving speedup. The total number of serial calls to $\mathcal{M}_{\theta}^h$ is the number of times the model processes a block of tokens. A block efficiency of 1 means that one is achieving no speedup relative to standard decoding.

% \begin{table}[t] \scriptsize \addtolength{\tabcolsep}{1.2pt}
% \begin{wraptable}{r}{0.43\textwidth}\scriptsize
% \vspace{-13pt}
% \centering
% \caption{Per-task test performance of each finetuned model and block efficiency.
% % Tasks include Language Modeling (LM), extractive Question Answering (QA), and both Long and Short Summarization (L-Sum \& S-Sum). 
% The performance metric for LM is perplexity, for QA is exact match, and for the remaining summarization tasks, the metric is ROUGE-L.}
% \label{tab:baseline}
% \addtolength{\tabcolsep}{-.5pt}
% \begin{tabular}{@{}c|l|c|c@{}} 
% \toprule
%  \multirow{2}{*}{Task} & \multirow{2}{*}{Dataset} & \multirow{2}{*}{Performance} & Block \\ 
%  & & & Efficiency \\\midrule
% LM                   & LAMBADA \cite{lambada}  & 7.88 & 3.12 \\ \midrule
% QA                   & SQuAD V1 \cite{squad}       & 57.60  & 2.08 \\ \midrule
% \multirow{2}{*}{S-SUM} & CNN/Daily \cite{cnn_sum}     & 39.85  & 1.74 \\
%                      & SAMSUM \cite{SAMSum}        & 37.66  & 1.27 \\ \midrule
% \multirow{3}{*}{L-SUM} & MultiNews \cite{multi_news}     & 23.08  & 1.10 \\
%                      & XSUM \cite{xsum}          & 52.15  & 1.13 \\ 
%                      & NewsRoom \cite{newsroom}      & 39.85  & 1.08 \\\bottomrule
% \end{tabular}
% \vspace{-20pt}
% \end{wraptable}

In addition to a standard language modeling dataset, LAMBADA \cite{lambada}, we conduct experiments across several classes of downstream tasks. In the realm of text summarization, we evaluate models on the XSUM \cite{xsum}, MultiNews \cite{multi_news}, SAMSum \cite{SAMSum}, NewsRoom \cite{newsroom} and CNN/DailyMail \cite{cnn_sum} datasets. Each of these datasets is characterized by distinct summary lengths and styles. For extractive QA, the SQuAD V1 dataset \cite{squad} serves as our testbed. For each task aside from language modeling, we finetune the blockwise parallel LM for that task.\footnote{Details are given in \textbf{\Autoref{sec:app_architecture}}.} \Tableautoref{tab:baseline}\footnote{The performance metric for LM is perplexity, for QA is exact match, and for the remaining summarization tasks, the metric is ROUGE-L.} shows that block efficiency varies dramatically across tasks and as a function of the number of block draft heads. We use all 9 block draft heads for subsequent experiments as this acts as an upper bound on possible block efficiency.

\Tableautoref{tab:case} sketches how BPD acts on three examples from each class of tasks.
\vspace{-5pt}
\begin{itemize}
    \item \textbf{LM:} BPD excels at generating common multi-word expressions in a single step. For example, (no) `thing more than', and (take) `his word for the' are each drafted and accepted in a single step.
    \item \textbf{QA:} BPD also attains high block efficiency in extractive QA, where it correctly drafts multi-token entities copied from the input sequence. In SQuAD V1, it accurately completes the answer `Grumman' from `Gru' by adding `mman', highlighting its ability to process multiple tokens at once and quickly extend answers.
    \item \textbf{SUM:} BPD's effectiveness in SUM tasks varies by dataset. For formulaic summaries like CNN/DailyMail, it performs well, reflecting its alignment with LM and QA tasks. However, in narrative-driven datasets like SAMSum and XSUM, where concise summaries are required, the block efficiency of BPD is little better than standard decoding.
\end{itemize}

\vspace{-15pt}
\section{Exploration of predictive dynamics in BPD}\label{sec:explore}
\vspace{-7pt}

\subsection{Consecutive repetition}
\label{subsec:consec_repetition}

\begin{wraptable}{r}{0.47\textwidth}\tiny
% \vspace{-12pt}
\centering
\caption{Consecutive token repetition in block drafts before and after C4-trained 2-gram rescoring of the top-16 lattice. ``\% Consec" is the percentage of consecutive identical draft tokens out of all pairs of consecutive tokens. ``Max run" is the average maximum repeated subsequence length in tokens (upper bound of 9, the number of block draft heads). Higher values correspond to more egregious repetition in drafts.}
\addtolength{\tabcolsep}{-.5pt}
\begin{tabular}{c|c|c|c|c|c}
    \toprule
         \multirow{2}{*}{Task} & \multirow{2}{*}{Dataset}  & \multicolumn{2}{c|}{\textbf{\% Consec}} & \multicolumn{2}{c}{\textbf{Max run}} \\\cmidrule{3-6}
         & & Vanilla & 2-gram & Vanilla & 2-gram \\ \midrule
         LM & LAMBADA & 20.0 & \underline{\textbf{10.7}} & 2.2 & \underline{\textbf{1.8}} \\ \midrule
         QA & SQuAD V1 & 75.5 & \underline{\textbf{67.6}} & 6.6 & \underline{\textbf{6.1}} \\ \midrule
         \multirow{2}{*}{S-SUM} & CNN/Daily & 46.4 & \underline{\textbf{21.9}} & 3.8 & \underline{\textbf{2.5}} \\
         & SAMSUM & 29.9 & \underline{\textbf{20.0}} & 3.1 & \underline{\textbf{2.5}} \\\midrule
         \multirow{3}{*}{L-SUM} & MultiNews & 33.6 & \underline{\textbf{14.7}} & 3.1 & \underline{\textbf{2.1}} \\
         & XSUM & 24.0 & \underline{\textbf{9.4}} & 2.6 & \underline{\textbf{1.7}} \\
         & NewsRoom & 47.2 & \underline{\textbf{32.1}} & 4.1 & \underline{\textbf{3.3}} \\\bottomrule
    \end{tabular}
    \vspace{-10pt}
    \label{tab:repetition_baseline}
\end{wraptable}

% \begin{table}[t]
%     \centering\small \addtolength{\tabcolsep}{-1.5pt}
%     \caption{Consecutive token repetition present in BPD drafts before and after C4-trained 2-gram rescoring of the top-16 lattice. ``\% Consec" is the percentage of consecutive identical draft tokens out of all pairs of consecutive tokens. ``Max Run" is the average maximum repeated subsequence length in tokens (upper bound of nine, the number of BPD heads). Higher values correspond to more egregious repetition in drafts.}
%     \begin{tabular}{c|c|c|c|c|c}
%     \toprule
%          \multirow{2}{*}{Task} & \multirow{2}{*}{Dataset}  & \multicolumn{2}{c|}{\textbf{\% Consec}} & \multicolumn{2}{c}{\textbf{Max Run}} \\\cmidrule{3-6}
%          & & Vanilla & 2-gram & Vanilla & 2-gram \\ \midrule
%          LM & LAMBADA & 20.0 & \underline{\textbf{10.7}} & 2.2 & \underline{\textbf{1.8}} \\ \midrule
%          QA & SQuAD V1 & 75.5 & \underline{\textbf{67.6}} & 6.6 & \underline{\textbf{6.1}} \\ \midrule
%          \multirow{2}{*}{S-SUM} & CNN/Daily & 46.4 & \underline{\textbf{21.9}} & 3.8 & \underline{\textbf{2.5}} \\
%          & SAMSUM & 29.9 & \underline{\textbf{20.0}} & 3.1 & \underline{\textbf{2.5}} \\\midrule
%          \multirow{3}{*}{L-SUM} & MultiNews & 33.6 & \underline{\textbf{14.7}} & 3.1 & \underline{\textbf{2.1}} \\
%          & XSUM & 24.0 & \underline{\textbf{9.4}} & 2.6 & \underline{\textbf{1.7}} \\
%          & NewsRoom & 47.2 & \underline{\textbf{32.1}} & 4.1 & \underline{\textbf{3.3}} \\\bottomrule
%     \end{tabular}
%     \vspace{-10pt}
%     \label{tab:repetition_baseline}
% \end{table}

We observe that vanilla block drafts are prone to significant token repetition. This is due to the fact that each head's prediction is independent of the others, and is a limitation shared with non-autoregressive generation in general \cite{nonar}. \autoref{tab:repetition_baseline} shows the proportion of consecutive tokens in block drafts that are identical to each other, along with the average maximum length of repeated sequences in block drafts across all decode time steps. We compare these statistics before and after rescoring with a 2-gram LM: a trivial rescorer, but one that can encourage local consistency between consecutive draft tokens. Strings of repeated tokens are unnatural, and unlikely to be generated by a strong base language model. Rescoring the top-$k$ lattice with even a simple language model eliminates a significant amount of repetition, reducing the percentage of consecutive repeated tokens from between \textbf{9.9\%} to \textbf{24.5\%}, depending on the task.
% \theertha{Percentage is relative, so I am not sure what it means by absolute. Consider just having reducing the percentage of consecutive repeated tokens from between \textbf{9.9\%} to \textbf{24.5\%}, depending on the task.}

\begin{figure*}
    \centering
    \begin{subfigure}{0.43\textwidth}
        \vspace{-10pt}
        \includegraphics[width=\linewidth]{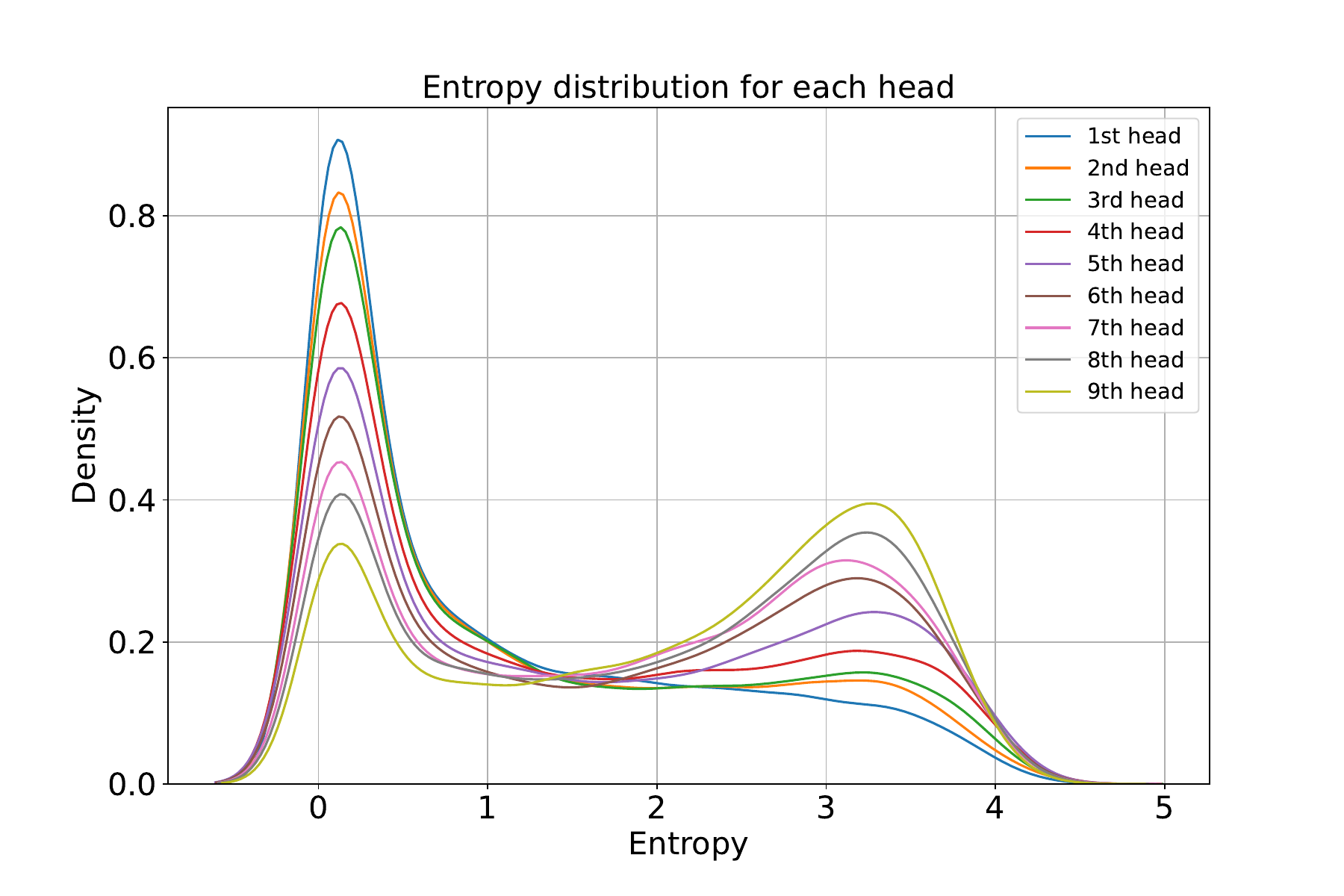}
        \vspace{-12.5pt}
        \caption{Entropy distributions across block draft heads}     \label{fig:entropy}
    \end{subfigure}
    \begin{subfigure}{0.48\textwidth}
        \includegraphics[width=\linewidth]{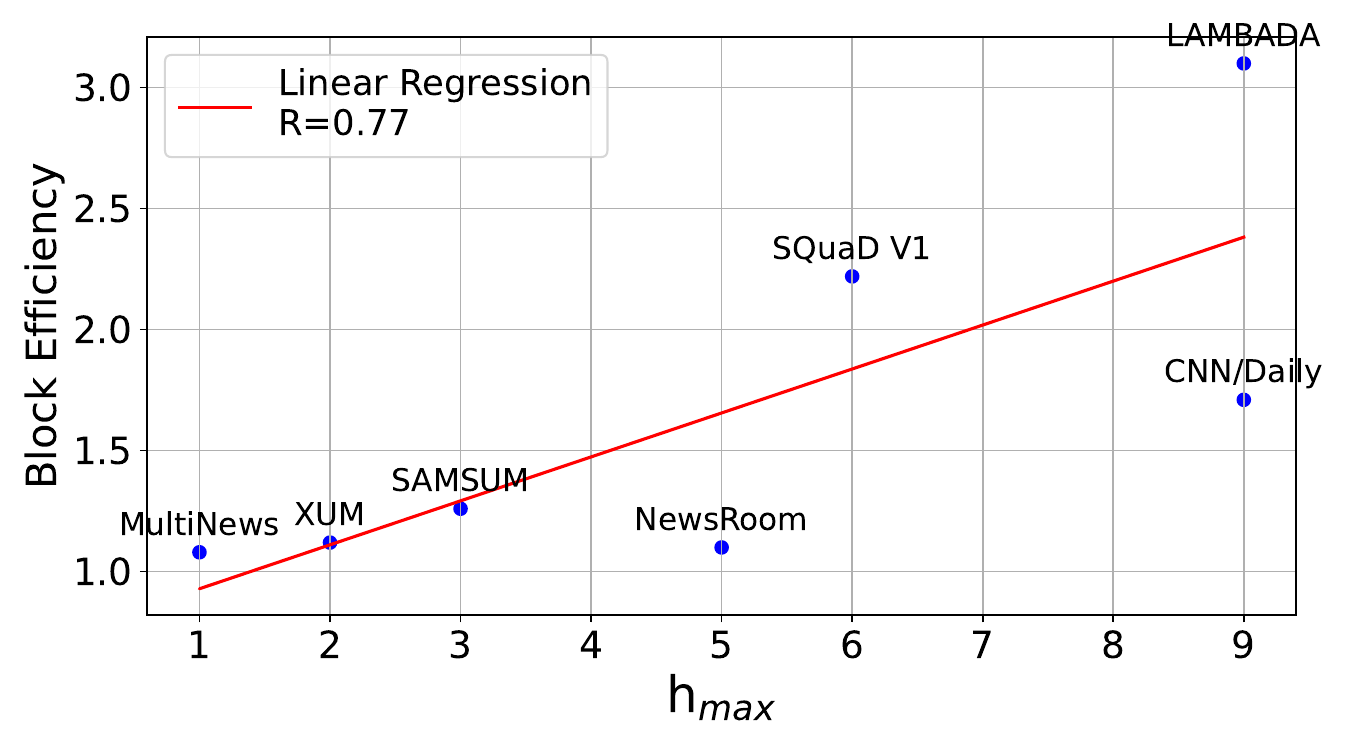}
        \vspace{-15pt}
        \caption{Correlation between block efficiency and $h_{\max}$} \label{fig:entropy_block}
    \end{subfigure}
    \vspace{-2pt}
    \caption{(a) Entropy distributions across block draft heads on LAMBADA. The density plots illustrate the entropy distribution for each head in the model. (b) Correlation between block efficiency and $h_{\max}$, the head until which the average entropy in a task increases monotonically.}
    \vspace{-10pt}
\end{figure*}

\vspace{-5pt}
\subsection{Confidence across multiple heads} \label{sec:cascade}
\vspace{-5pt}

Intuitively, predicting the identity of the $i^\text{th}$ future token becomes harder as $i$ increases. To better understand this phenomenon, we measure the confidence of the predictions by the entropy of the probability distribution.  In \Autoref{fig:entropy}, we plot the normalized histogram of entropy of each head on the LAMBADA task. From the normalized histogram, it is clear that the entropy increases as we move from first head to the last head, which agrees with our intuition that hardness of predictions increases as $i$ increases. 

However, we observed that the entropy of heads does not increase monotonically for all tasks. Let \(\overline{\mathbb{H}}[i]\) be the average entropy of head $i$ on a particular corpus, and let \(h_{\max} = \max\limits_{k} \{k: \forall i < k, \overline{\mathbb{H}}[i] \leq \overline{\mathbb{H}}[i+1] \}\), be the index of the largest head such that the average entropy of each head increases monotonically to that point. We observed a strong correlation between $h_{\max}$ and block efficiency (\Autoref{fig:entropy_block}). Heads with lower entropy (indicating more confident predictions) intuitively contribute more to efficiency. Nonetheless, simply maximizing the number of low-entropy heads is not optimal, but rather incorporating progressively higher entropy heads, up to a certain point, can benefit decoding efficiency. A linear regression confirms this with an R-value of $\mathbf{0.77}$. This analysis suggests that BPD head entropy could be used as a proxy for block efficiency, and thus inference latency.

\vspace{-5pt}
\subsection{Oracle top-\textit{k} block efficiency}\label{sec:oracle}
\vspace{-5pt}
\begin{figure*}[t!]
\centering
\includegraphics[width=\textwidth]{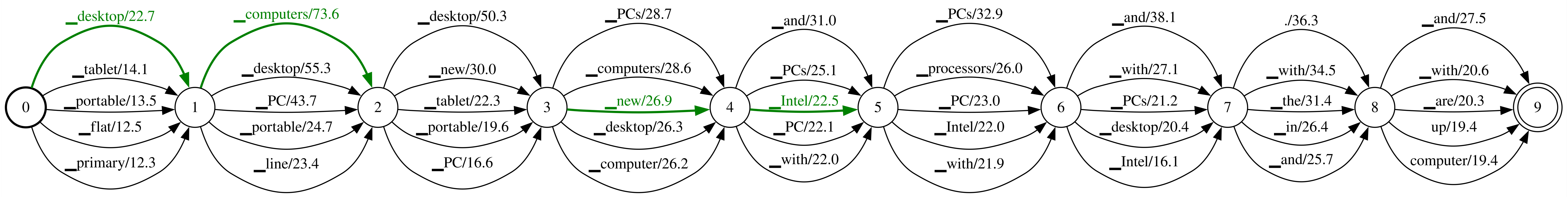}
%%\vspace{-10pt}
\caption{An example of a top-5 sausage lattice generated on a NewsRoom example. Edge weights correspond to (rescored) logits. Edges at each time step are ordered in descending weight and green, bolded edges correspond to candidates matching the greedy decode over the next nine tokens: \textit{"... desktop computers with new Intel Corp processors that it ..."}. The initial node in this graph is state 0 and the final node is 9.}
\vspace{-10pt}
\label{fig:example_sausage_lattice}
\end{figure*}

% \begin{figure}[t]
% \centering
% \includegraphics[width=\columnwidth]{arxiv_figs/oracle_lattice.pdf}
% \vspace{-20pt}
% \caption{Illustration of the output through oracle selection. For a given top $k$ of 3, if we can choose the oracle path successfully, the block efficiency can be improved from 1 to 5. }
% \vspace{-15pt}
% \label{fig:oracle_lattice}
% \end{figure}

\textbf{Oracle efficiency}\, The concept of oracle block efficiency in BPD serves as a theoretical benchmark, illustrating the headroom available from improving the quality of the block draft. To compute oracle block efficiency, we consider the top-$k$ most probable tokens at each head, and form a ``sausage'' lattice from these. This data structure is a weighted directed graph, which succinctly represents all possible drafts (and their score under the BPD model) that could be formed from selecting one of $k$ tokens from each of the $h$ heads (\autoref{fig:example_sausage_lattice}). In the automatic speech recognition and machine translation communities, it is known as a ``confusion network'' \cite{jhuws2006iwslt,moses}.

\begin{wrapfigure}{r}{0.45\textwidth}
    \vspace{-12pt}
    \includegraphics[width=1.0\linewidth]{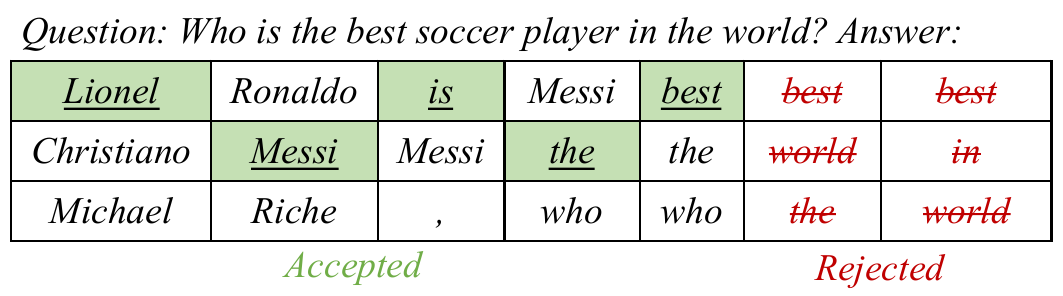}
    \vspace{-15pt}
    \caption{Illustration of the output through oracle selection. For a given top $k$ of 3, if we can choose the oracle path successfully, the block efficiency can be improved from 1 to 5. } 
    \vspace{-15pt}
    \label{fig:oracle_lattice}
\end{wrapfigure}

Given the top-$k$ lattice at each decoding step, we identify an \textit{oracle path} that represents the path through the lattice that maximizes the length of the accepted prefix. This exercise, as shown in \autoref{fig:oracle_lattice}, gives us insight into how much headroom exists in improving block drafts.

\begin{figure*}
    \centering
    \begin{subfigure}{0.32\textwidth}
        \includegraphics[width=\linewidth]{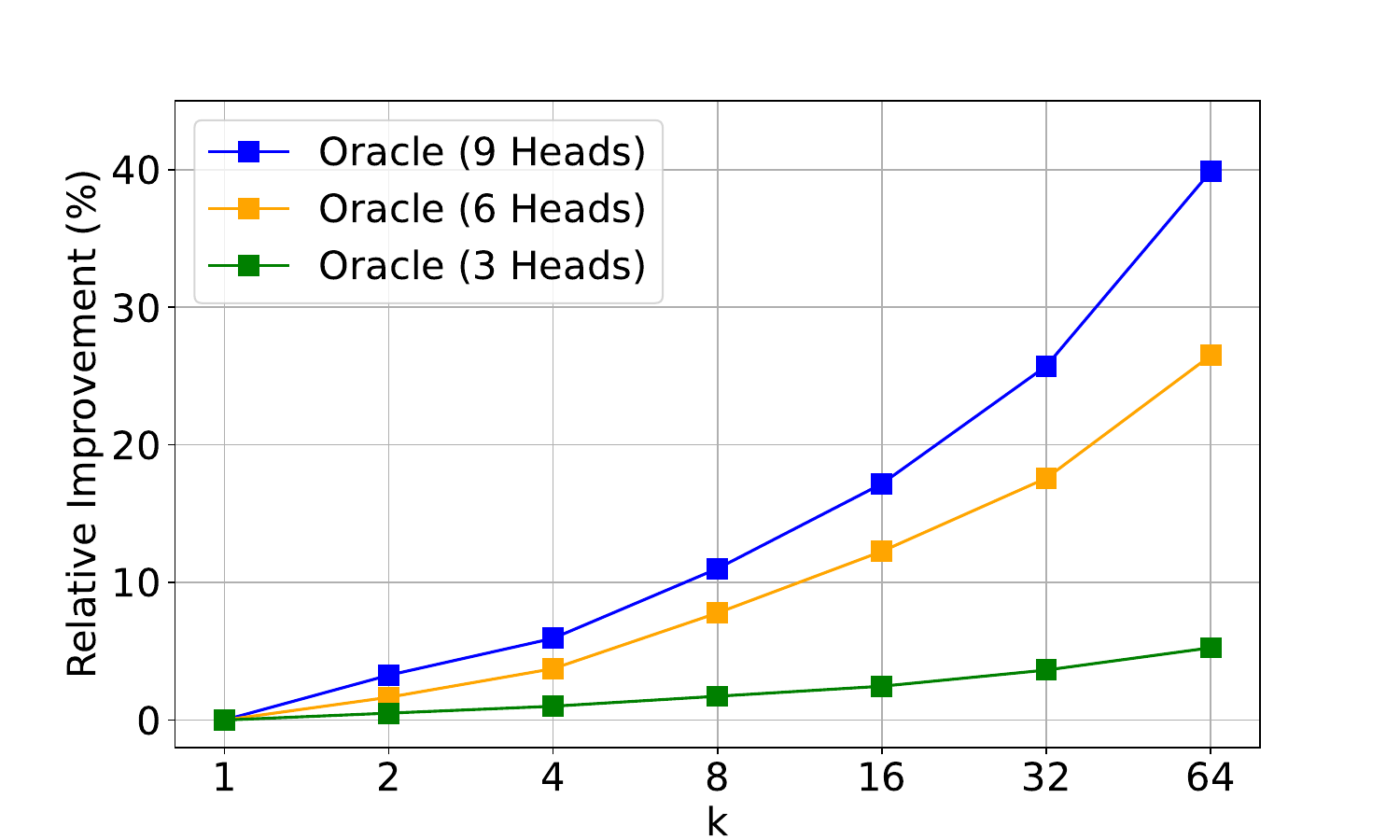}
        % \vspace{-15pt}
        \caption{LAMBADA}
    \end{subfigure}
    \begin{subfigure}{0.32\textwidth}
        \includegraphics[width=\linewidth]{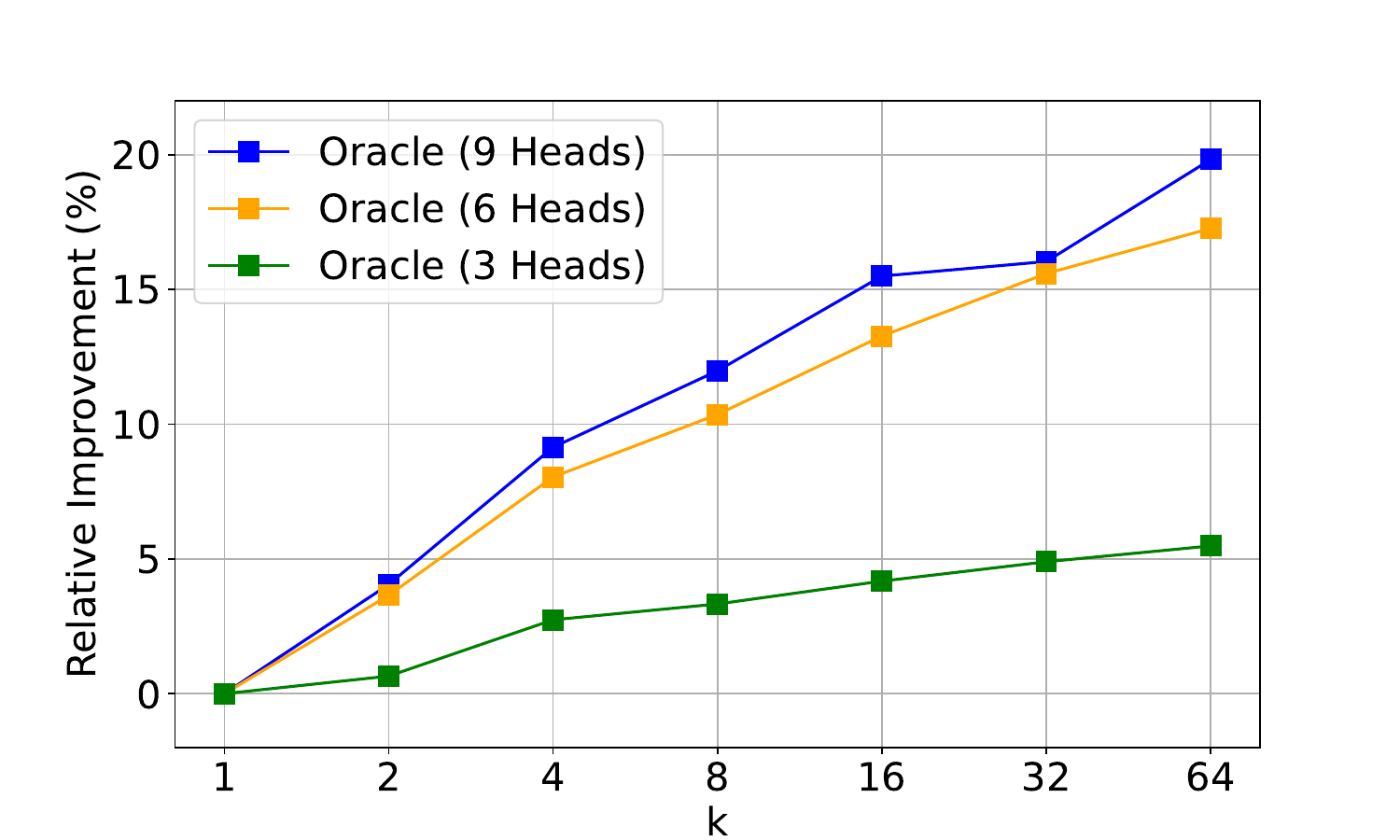}
        % \vspace{-15pt}
        \caption{SQuAD V1}
    \end{subfigure}
    \begin{subfigure}{0.32\textwidth}
        \includegraphics[width=\linewidth]{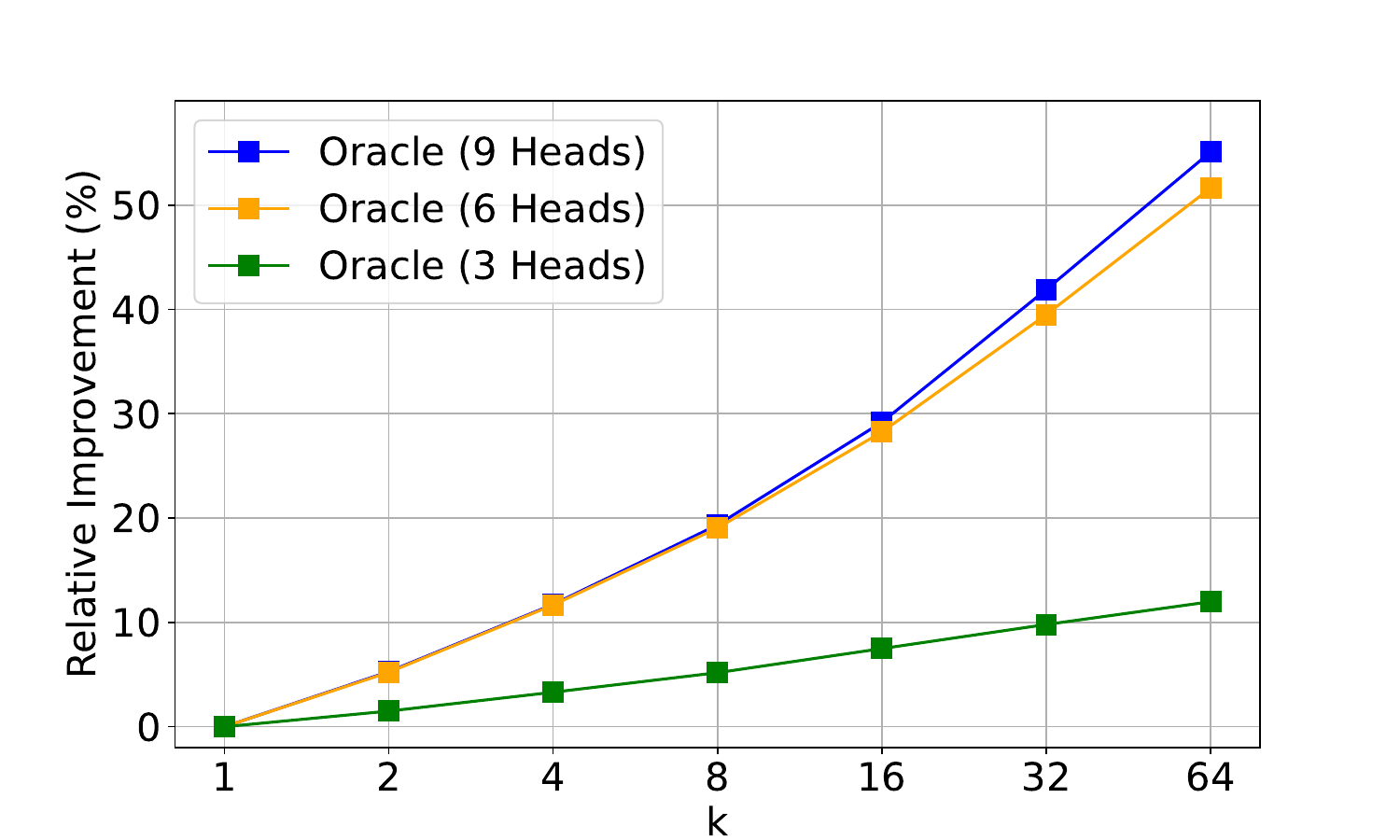}
        % \vspace{-15pt}
        \caption{CNN/Daily}
    \end{subfigure}
    \newline
    \begin{subfigure}{0.32\textwidth}
        \includegraphics[width=\linewidth]{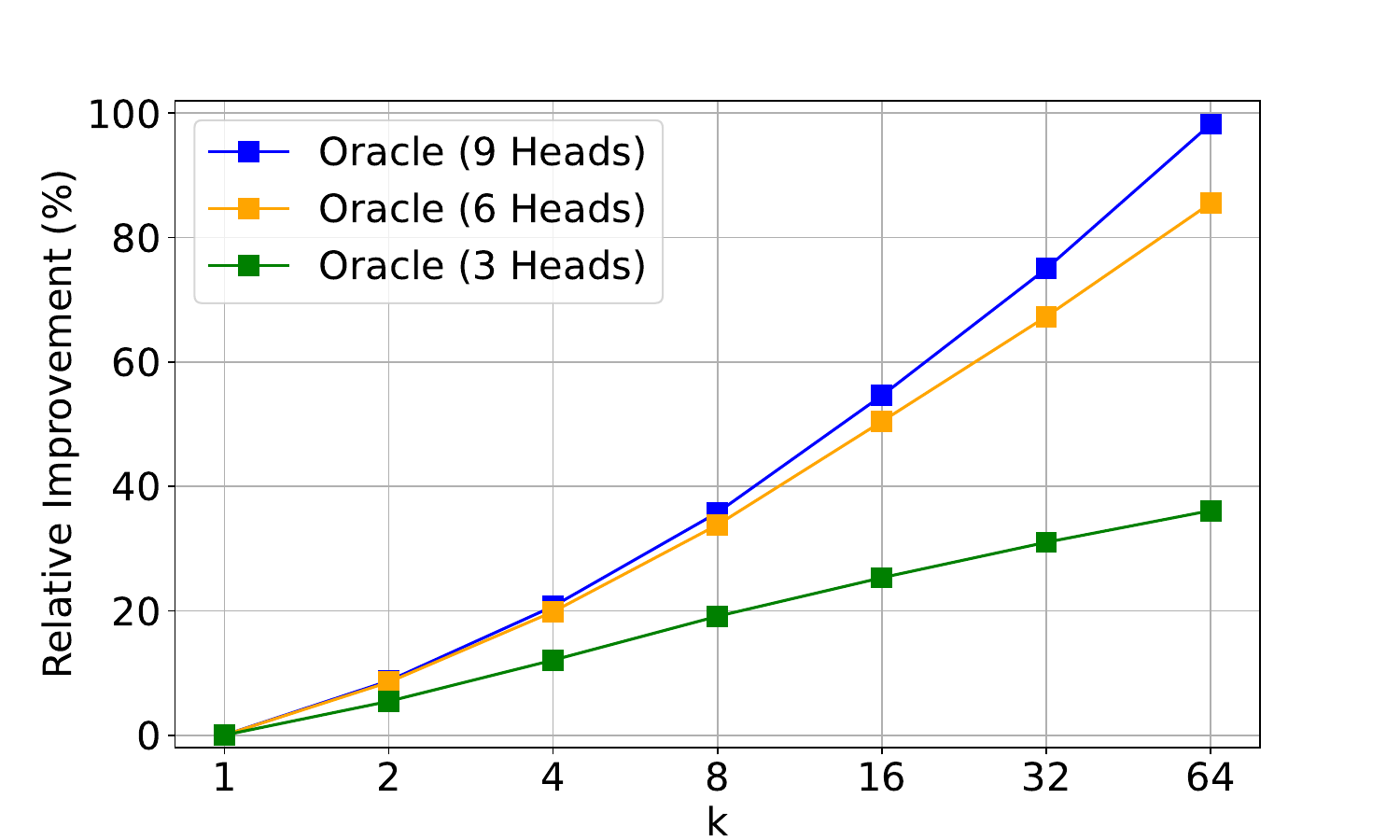}
        % \vspace{-15pt}
        \caption{SAMSUM}
    \end{subfigure}
    \begin{subfigure}{0.32\textwidth}
        \includegraphics[width=\linewidth]{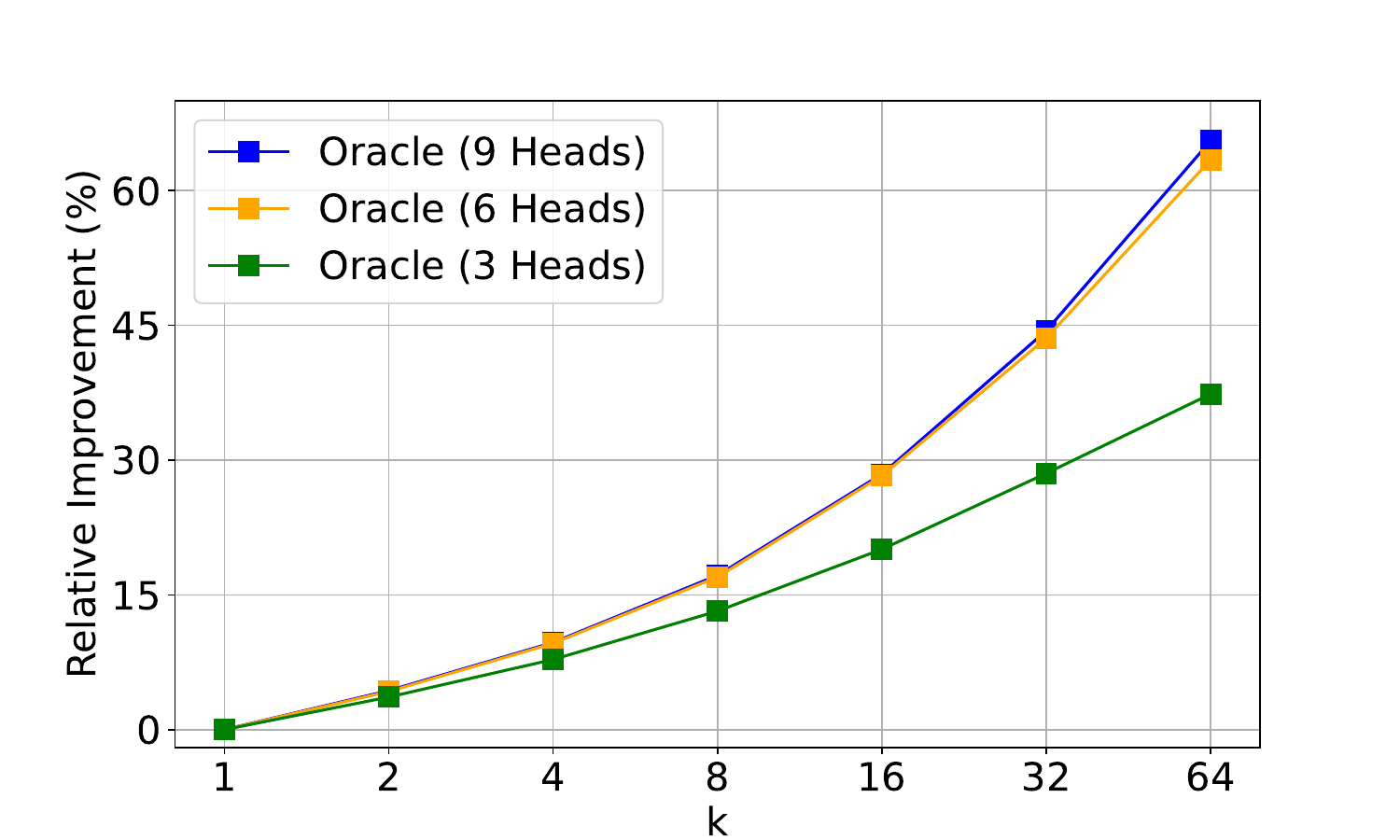}
        % \vspace{-15pt}
        \caption{MultiNews}
    \end{subfigure}
    \begin{subfigure}{0.32\textwidth}
        \includegraphics[width=\linewidth]{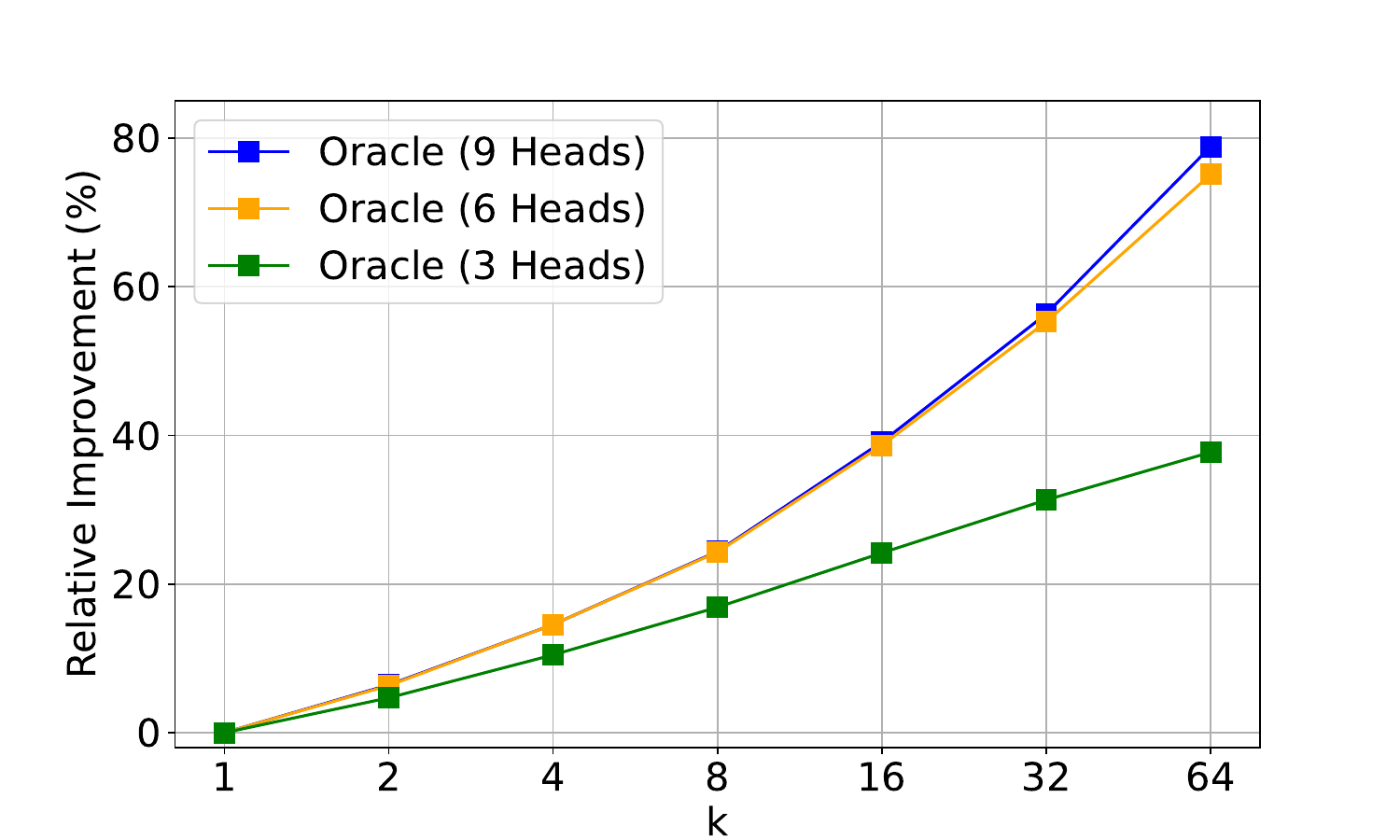}
        % \vspace{-15pt}
        \caption{XSUM}
    \end{subfigure}
    \caption{Oracle block efficiency over the top-$k$ lattice as a function $k$.    Each plot (a-f) represents a different task, demonstrating the relative improvement in block efficiency of the oracle draft with respect to the standard block draft as a function of the number of block draft heads used.
    }
    \vspace{-5pt}
    \label{fig:oracle}
\end{figure*}

\textbf{Potential headroom from oracle selection}\, Oracle drafting is not practical, but rather a reference point. Analyzing the gap between actual BPD performance and the oracle upper bound (\autoref{fig:oracle}) helps us to understand the limitations of the original block drafts and potential areas for improvement. Additionally, exploring oracle efficiency as a function of the $k$ in the top-$k$ lattice, demonstrates how ``close'' the block draft was to producing a stronger draft.

\vspace{-10pt}
\section{Lattice rescoring for improved block efficiency}
\vspace{-5pt}

Having explored BPD's prediction dynamics, we propose two drafting algorithms to improve block efficiency through rescoring of the top-$k$ lattice. This section presents techniques for rescoring the top-$k$ lattice along with empirical results.

Each of these algorithms is a modification of the block drafted in \textbf{Stage 1} in Algorithm~\ref{alg:bpd}. Instead of using the most likely token at each head as the prediction, we construct the top-$k$ sausage lattice of likely drafts from each head, where the set of top-$k$ tokens is denoted as $S_i$ for head $i$. This approach allows any token within $S_i$ to be chosen for position $i$, yielding a total possible combinations of:

\vspace{-10pt}
\[
|S_1| \times |S_2| \times \ldots |S_h| = k^h
\]
%We refer to the set of all these possible combinations as the \emph{lattice}. \Autoref{fig:example_sausage_lattice} gives an example of a top-5 BPD sausage lattice. 
% Observe that any path from the state $0$ to state $9$ is a path possible by the algorithm.

In this lattice, any path from the start to final state represents a viable draft. Two algorithms are proposed to select a small number of $h$-length drafts from this lattice, which are then passed to the verification step. The first algorithm employs neural autoregressive transformers (\textbf{\Autoref{sec:neural}}), while the second utilizes \ngram language models (\textbf{\Autoref{sec:ngram}}).

\begin{table*}[t]    \centering \small % \addtolength{\tabcolsep}{-0.5pt}
    %%\vspace{-5pt}
    \caption{Block efficiency of our methods over the top-$16$ lattice. `16-best 0-gram BPD' indicates performance of 16-best verification over the original lattice without \ngram{} rescoring. Relative percent improvement over BPD (Baseline) is indicated in parentheses. Green circles (\textcolor{green!60!black}{$\CIRCLE$}) indicate improvement over the Baseline, while red circles (\textcolor{red!60!black}{$\CIRCLE$}) denote no improvement.
    % \taehyeon{@Taehyeon: looks better, how about replacing (new) with rescoring as follows?}
    % \theertha{Theertha:  I edited the table a bit by adding the local and global columns, please take a look. if it looks okay, please propagate it to other tables (e.g., table 9).  }
    }
    \resizebox{\textwidth}{!}{%
    \begin{tabular}
    {c|c|c|c|c|c|c|c} 
    \toprule
        \multirow{2}{*}{Task} & \multirow{2}{*}{Dataset}   & Baseline & Local rescoring & \multicolumn{3}{c|}{Global rescoring} &  \multirow{2}{*}{Oracle (k=16)}   \\
          &   & BPD & neural-61M BPD & 4-gram BPD &  16-best 0-gram BPD & 16-best 4-gram BPD &  \\
         \midrule
        LM & LAMBADA   & {3.12} & 3.08 (-1.28\%) \textcolor{red!60!black}{$\CIRCLE$} & 3.05 (-2.24\%) \textcolor{red!60!black}{$\CIRCLE$} & 3.23 (+3.53\%) \textcolor{green!60!black}{$\CIRCLE$} & \underline{\textbf{3.29 (+5.45\%)}} \textcolor{green!60!black}{$\CIRCLE$}  & 3.67 \\ \midrule
        QA & SQuAD V1  & 2.08 & {2.10 (+0.96\%)} \textcolor{green!60!black}{$\CIRCLE$}& 2.07 (-0.48\%) \textcolor{red!60!black}{$\CIRCLE$}  & 2.18 (+4.85\%) \textcolor{green!60!black}{$\CIRCLE$} & \underline{\textbf{2.22 (+6.87\%)}} \textcolor{green!60!black}{$\CIRCLE$} & 2.45\\ \midrule
        \multirow{2}{*}{S-SUM} & CNN/Daily & {1.74}& 1.73 (-0.57\%) \textcolor{red!60!black}{$\CIRCLE$} &  1.73 (-0.57\%) \textcolor{red!60!black}{$\CIRCLE$}  & 1.82 (+4.66\%) \textcolor{green!60!black}{$\CIRCLE$} & \underline{\textbf{1.83 (+5.41\%)}} \textcolor{green!60!black}{$\CIRCLE$} & 2.26 \\ 
        & SAMSUM    & 1.27   & {1.39 (+9.45\%)} \textcolor{green!60!black}{$\CIRCLE$} & 1.29 (+1.57\%) \textcolor{green!60!black}{$\CIRCLE$} & 1.37 (+7.87\%) \textcolor{green!60!black}{$\CIRCLE$} & \underline{\textbf{1.45 (+14.17\%)}} \textcolor{green!60!black}{$\CIRCLE$} & 1.95 \\ \midrule
        \multirow{3}{*}{L-SUM} & MultiNews & 1.10 & \underline{\textbf{1.25 (+13.64\%)}} \textcolor{green!60!black}{$\CIRCLE$} & 1.12 (+1.82\%) \textcolor{green!60!black}{$\CIRCLE$} & 1.13 (+2.73\%) \textcolor{green!60!black}{$\CIRCLE$} & 1.22 (+10.91\%) \textcolor{green!60!black}{$\CIRCLE$} & 1.43 \\
        & XSUM      & 1.13 & {1.23 (+8.85\%)} \textcolor{green!60!black}{$\CIRCLE$}& 1.16 (+2.65\%) \textcolor{green!60!black}{$\CIRCLE$}  & 1.18 (+4.42\%) \textcolor{green!60!black}{$\CIRCLE$} & \underline{\textbf{1.26 (+11.50\%)}} \textcolor{green!60!black}{$\CIRCLE$} & 1.55\\
        & NewsRoom  & 1.08  & {1.29 (+19.44\%)} \textcolor{green!60!black}{$\CIRCLE$} & 1.18 (+9.26\%) \textcolor{green!60!black}{$\CIRCLE$} & 1.11 (+2.78\%) \textcolor{green!60!black}{$\CIRCLE$} & \textbf{\underline{1.31 (+21.30\%)}} \textcolor{green!60!black}{$\CIRCLE$} & 1.50 \\
    \bottomrule
    \end{tabular}
    }
    \vspace{-20pt}
    \label{tab:fst_results}
\end{table*}

\vspace{-5pt}
\subsection{Local rescoring via neural models}\label{sec:neural}
\vspace{-5pt}

\label{sec:algorithms}
\begin{wrapfigure}{r}{0.52\textwidth} \small
\vspace{-15pt}
\begin{algorithm}[H] 
\caption{Local rescoring via neural models}
    \label{algorithm} \label{alg:rescoring}
    \begin{algorithmic}[1]
    \INPUT: Blockwise parallel LM $\mathcal{M}_{\theta}^h$, top-$k$ indices selection function \textsc{Top-}$k(\cdot)$, rescoring model $\mathcal{M}_{\theta_r}$, interpolation weight $\alpha>0$.\\
    \STATE $z^1_{t+1}, \dots, z^h_{t+h} \leftarrow \mathcal{M}_{\theta}^h(y_{t+1}, \dots, y_{t+h} |\bar{x}, y_{\leq t})$ \\
    \STATE $S_1, \dots, S_h \leftarrow \textsc{Top-}k(z^1_{t+1}, \dots, z^h_{t+h})$ 
    \\
    \STATE \textcolor{blue}{\textbf{/* Lattice Rescoring (\Autoref{sec:neural})*/}} \\
    \FOR{$j \leftarrow 2, \dots, h$ in parallel}
    \STATE $r_{t+j} \leftarrow \mathcal{M}_{\theta_r}(y_{t+j} |x, y_{< t+j})$
    \STATE $z^j_{t,j}[S_j] \leftarrow z^j_{t,j}[S_j] + \alpha \cdot r_{t+j}[S_j]$
    \ENDFOR
\end{algorithmic}
\end{algorithm}
% \vspace{-10pt}
\end{wrapfigure}

A simple approach uses a small neural rescorer, interpolating between the logits of the rescorer LM and vanilla block draft logits with an interpolation weight (\textbf{\Autoref{alg:rescoring}}). The rescored prediction is given by:

\vspace{-15pt}
\begin{gather*}
    z^j_{t,j} [S_j] \leftarrow z^j_{t,j}[S_j] + \alpha \cdot r_{t+j}[S_j]
\end{gather*}

where $z^j_{t,j}$ represents the logit of the block draft at head $j$, and $r_{t+j}$ is the corresponding logit predicted by the small neural rescoring model, which is conditioned on the sequence \( y_{\leq t}, \cdots, \hat{y}_{t+2}, \cdots, \hat{y}_{t+j-1} \). The parameter \( \alpha \) is the weight placed on the rescorer's prediction. We experiment with decoder-only transformers having 32, 61, and 94 million (M) weight parameters (\textbf{\Autoref{sec:app_architecture}}). We use greedy rescoring when generating the neural draft.

\subsection{Global \ngram{}rescoring} \label{sec:ngram}
We also evaluate the quality of drafts generated by rescoring with an \ngram LM. 
Recall that blockwise parallel LMs can be used to compute a lattice representing $k^h$ possible sequences. We rescore all of these sequences with an \ngram{} model, select the top $p$ sequences and pass them to the verification stage. When $p = 1$, we refer to this as \emph{\ngram rescoring} and when $p > 1$, we refer to this as \emph{$p$-best \ngram BPD}.

While global rescoring typically yields better results compared to local rescoring, rescoring $k^h$ sequences with a neural LM and selecting the most likely sequence would take time $O(k^h)$, which is computationally prohibitive in most cases. 
Hence, we take advantage of \ngram models, which are unique in that we can select the most likely sequence in time $\text{poly}(k ,h)$ using dynamic programming. 
We use the OpenFST library \cite{openfst} to represent each \ngram model as a weighted finite state automaton and apply finite state composition with the top-$k$ lattice followed by extraction of the $p$ most likely draft sequences. Training details for the \ngram models are given in \textbf{\textcolor{blue}{Appendix}~\ref{sec:app_ngram_details}}.

% adrian: do you think its good to move the fst talk in Section 7.3 here? I think its a complete story.
% }

\begin{figure*}
    \centering
    \begin{subfigure}{0.32\textwidth}
        \includegraphics[width=\linewidth]{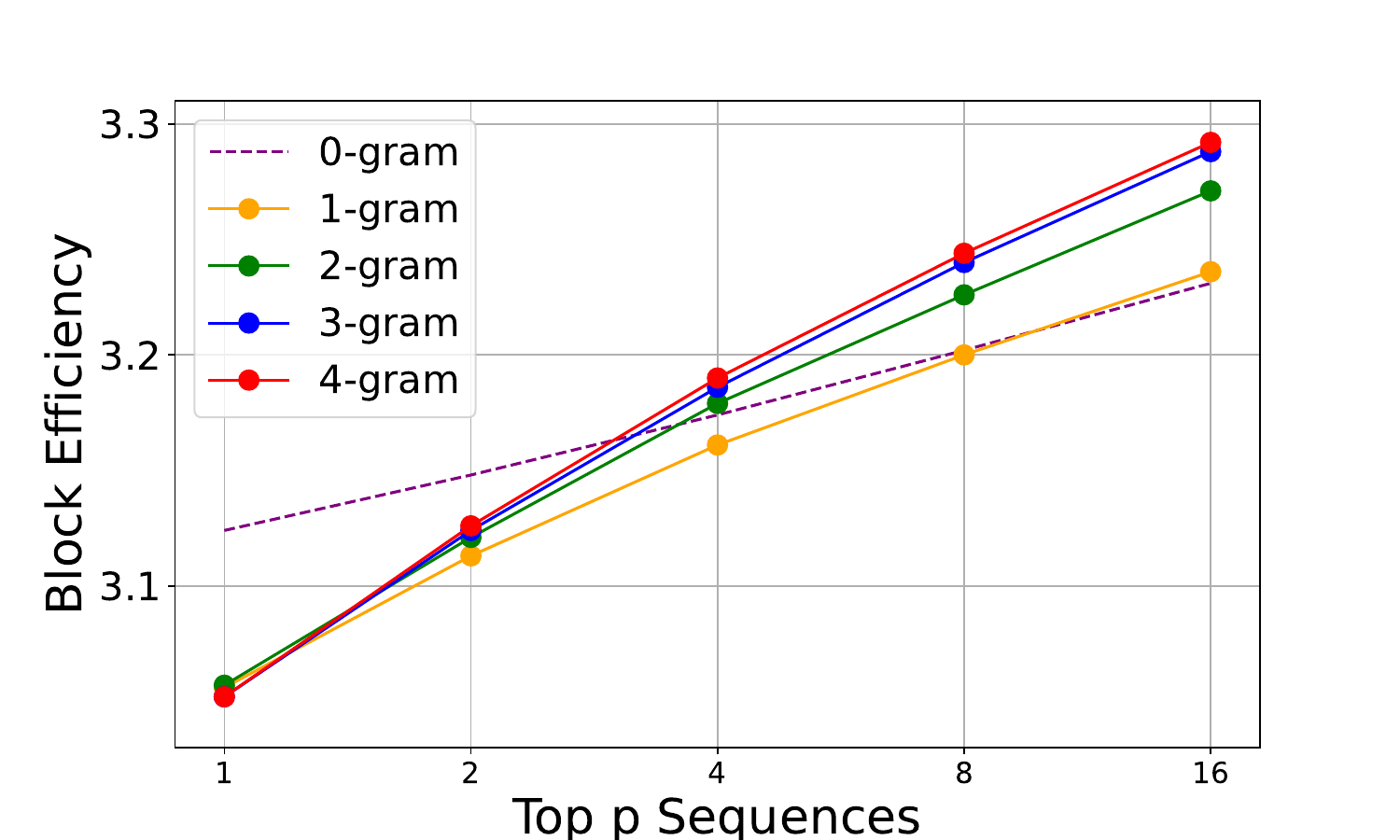}
        % \vspace{-15pt}
        \caption{LAMBADA}
    \end{subfigure}
    \begin{subfigure}{0.32\textwidth}
        \includegraphics[width=\linewidth]{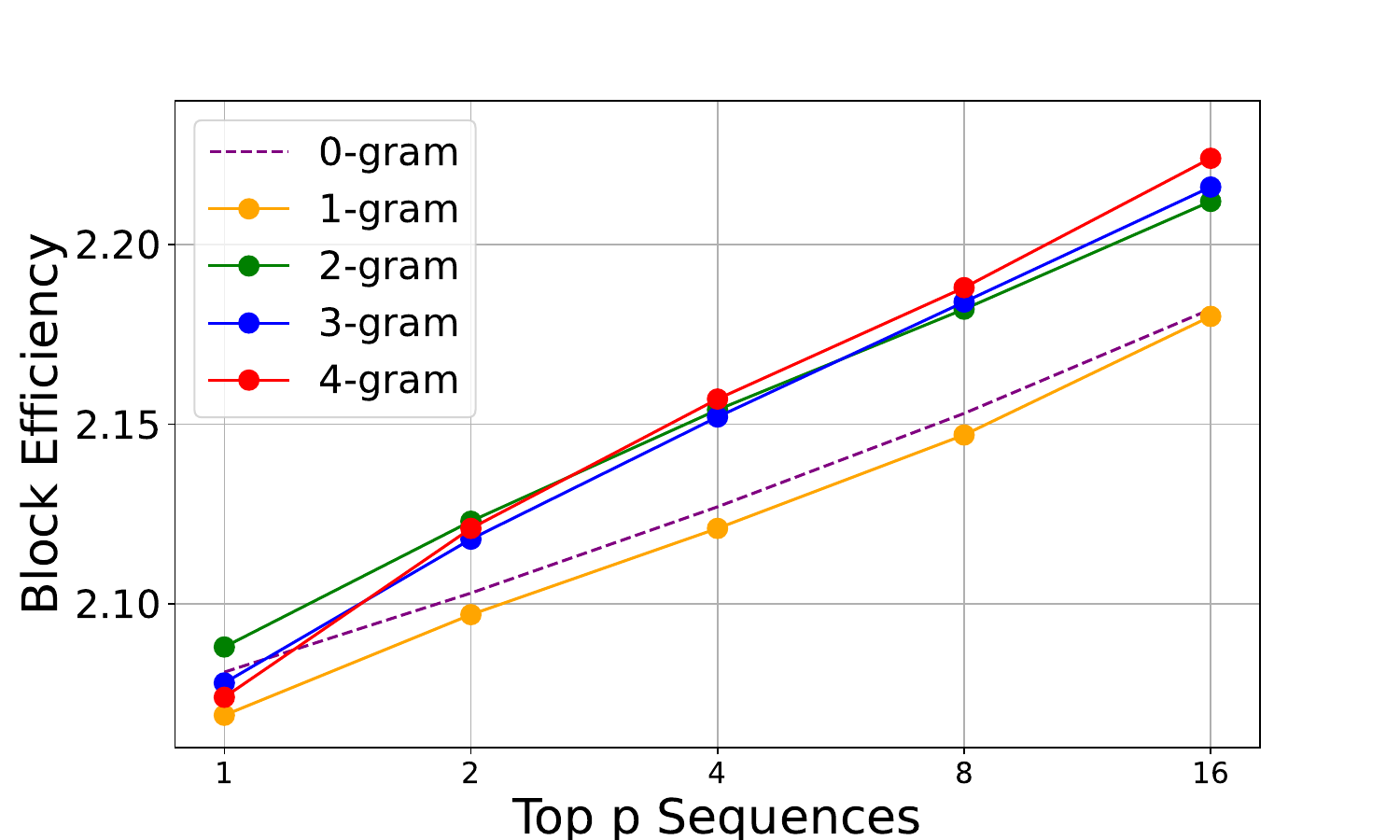}
        % \vspace{-15pt}
        \caption{SQuAD V1}
    \end{subfigure}
    \begin{subfigure}{0.32\textwidth}
        \includegraphics[width=\linewidth]{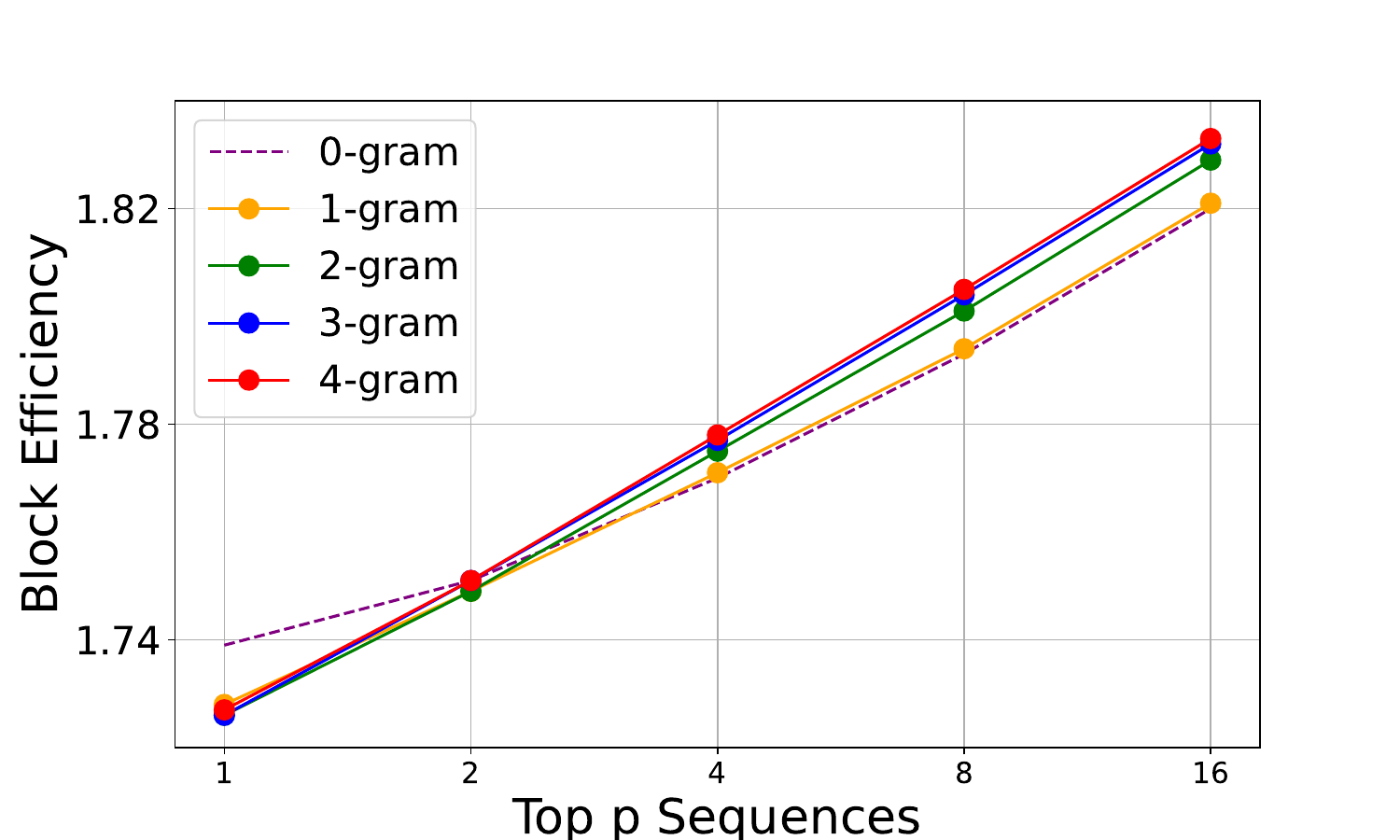}
        % \vspace{-15pt}
        \caption{CNN/Daily}
    \end{subfigure}
    \vspace{-2pt}
    \newline 
    \begin{subfigure}{0.32\textwidth}
        \includegraphics[width=\linewidth]{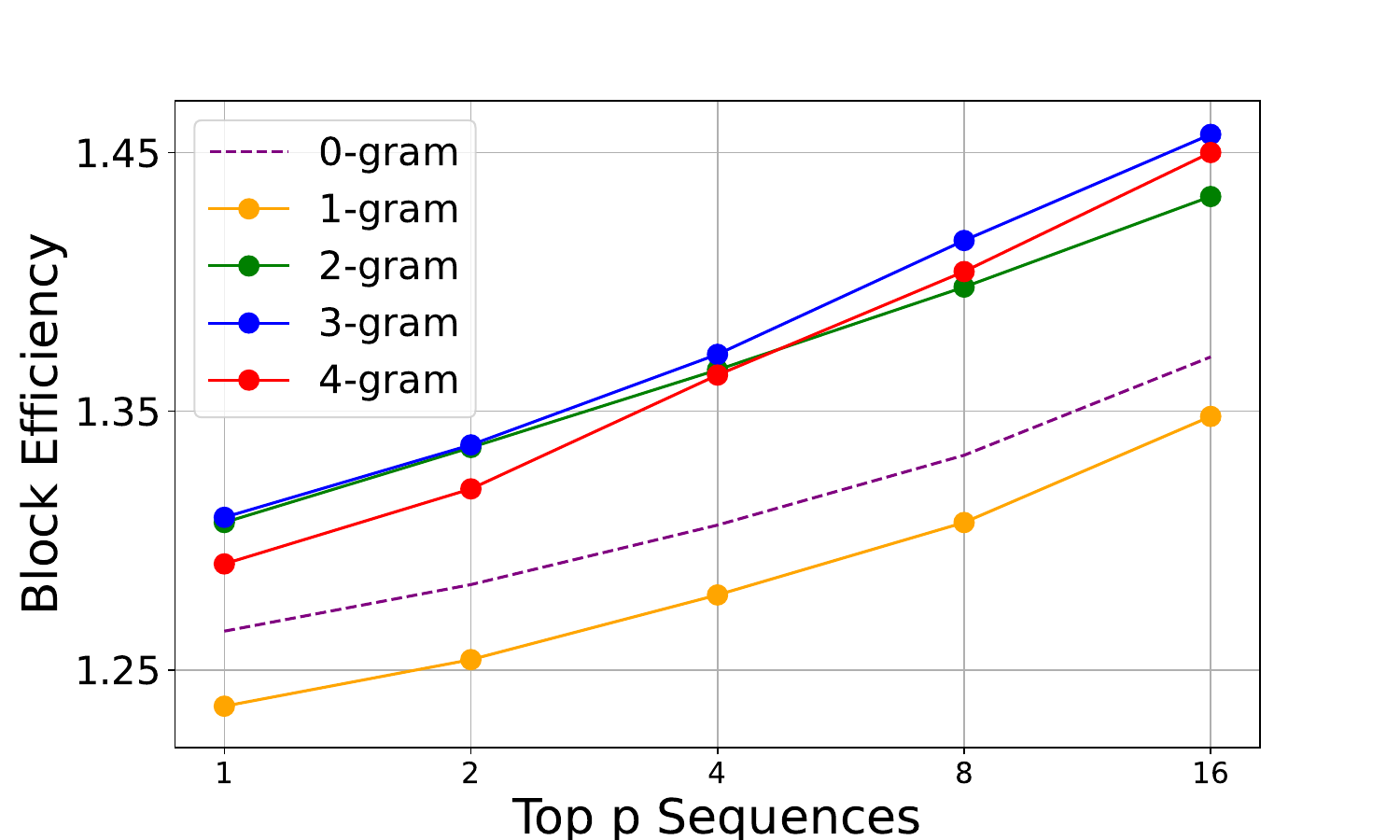}
        % \vspace{-15pt}
        \caption{SAMSUM}
    \end{subfigure}
    \begin{subfigure}{0.32\textwidth}
        \includegraphics[width=\linewidth]{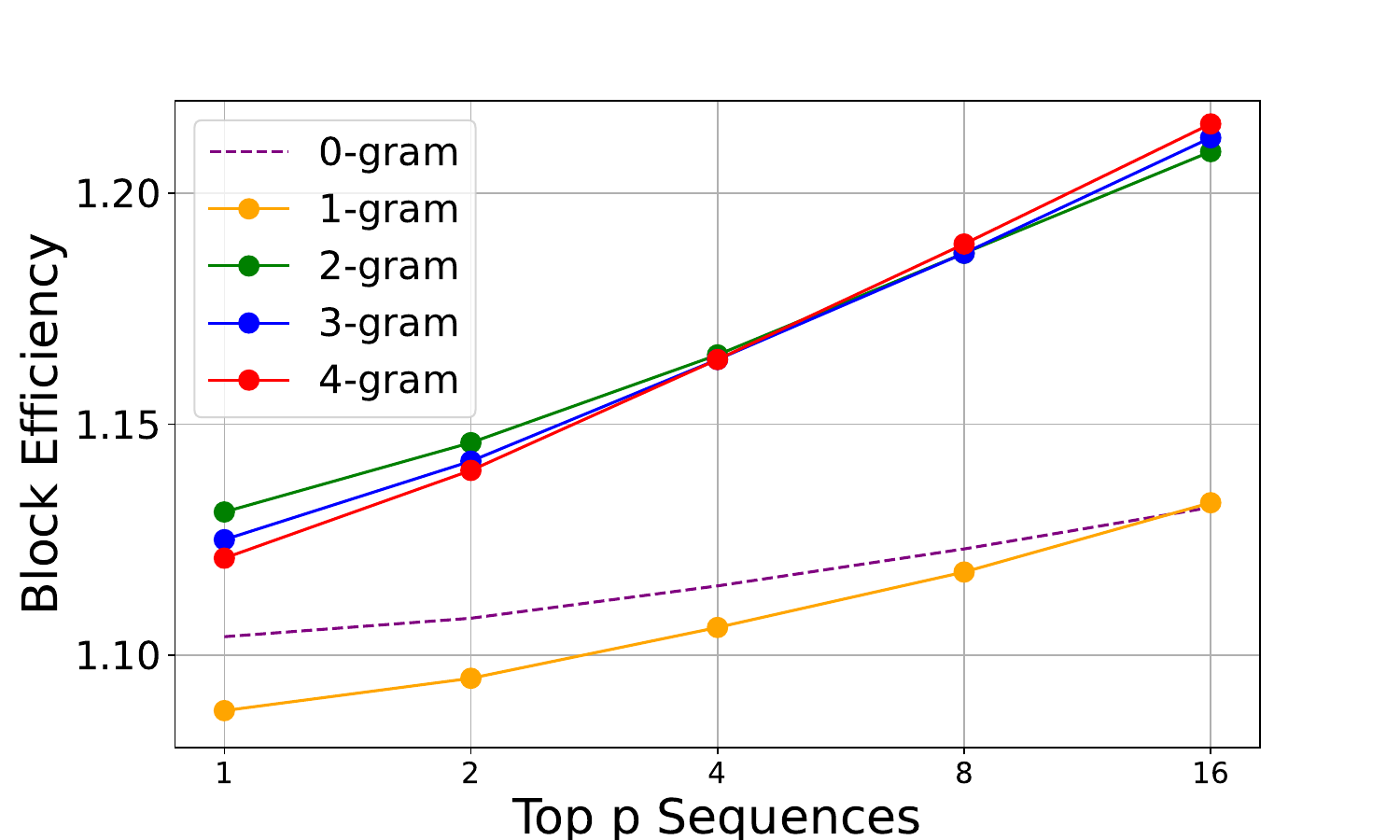}
        % \vspace{-15pt}
        \caption{MultiNews}
    \end{subfigure}
    \begin{subfigure}{0.32\textwidth}
        \includegraphics[width=\linewidth]{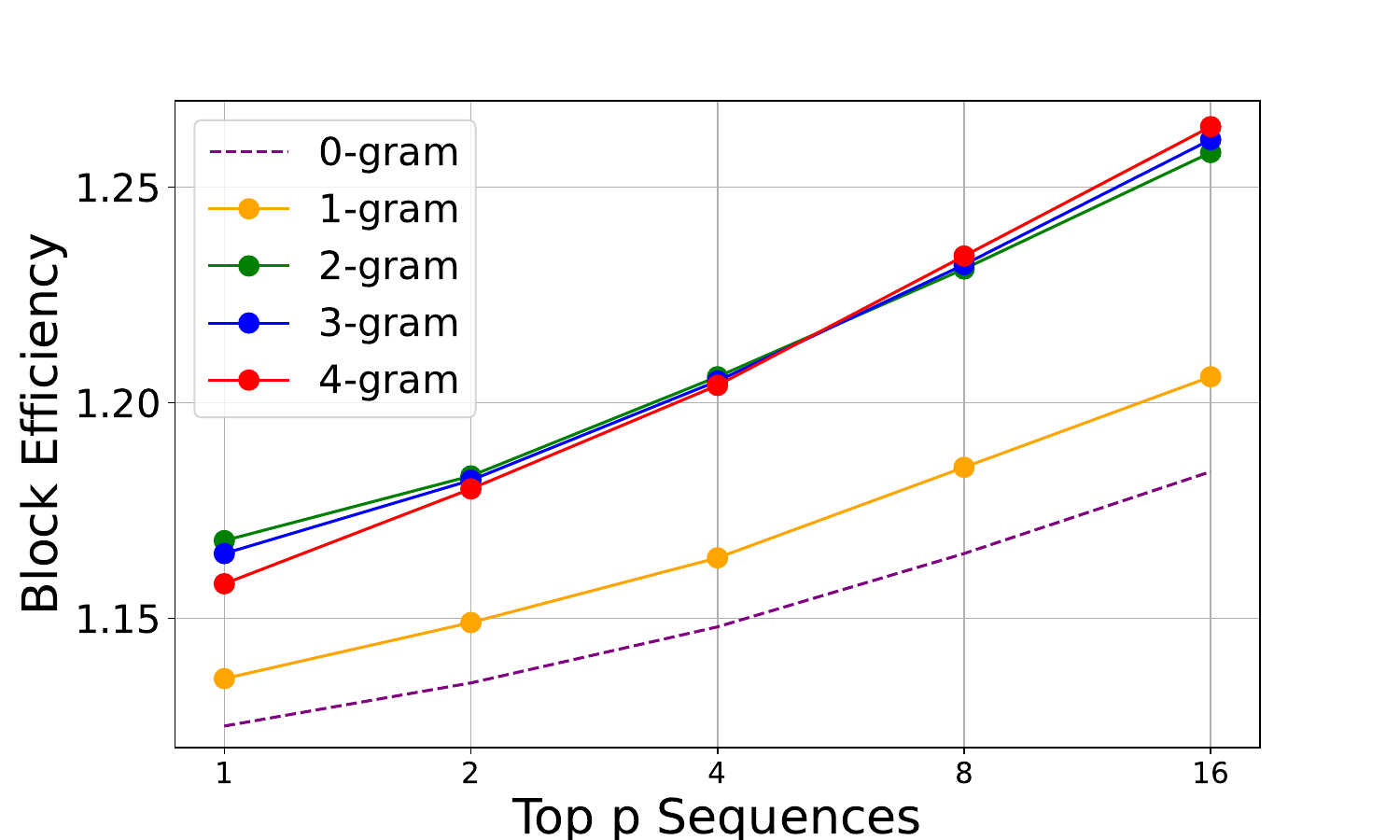}
        % \vspace{-15pt}
        \caption{XSUM}
    \end{subfigure}
    \vspace{-5pt}
    \caption{Block efficiency of $p$-best \ngram BPD methods as a function of the number of \textcolor{black}{top $p$ sequences} verified in parallel. The block efficiency of the methods is evaluated with the the same number of paths extracted from the top-$16$ lattice. 
    % \adrian{Not sure what the last sentence is trying to convey.} $0$-gram indicates $p$-best BPD without any rescoring and the number of \textcolor{black}{top $p$ sequences} is varied in ${1, 2, 4, 8, 16}$ over a top-$16$ lattice. 
    % \theertha{It might be good to change the label of x axis to $k$ or $top-k$ and be consistent with Figure 6.}
    }
    \label{fig:parallel_ngram_bpd}
    \vspace{-5pt}
\end{figure*}

% \adrian{TODO: cite FST for ASR decoding paper? move \ngram training details to appendix? shortest path under the tropical semiring.... Much of this will have to be moved to an appendix.}

% \adrian{We do tune interpolation weights for \ngram{} models. That bit is shared with the neural rescoring. I vote for including discussion around generating a $b$-best batch for neural rescoring is less straightforward: not globally optimal paths, need to implement via beam search, which may be cumbersome to implement efficiently.}

%\begin{figure*}[t]
%\centering
%\includegraphics[width=\textwidth]{arxiv_figs/wall_clock.pdf}
%\vspace{-25pt}
%\caption{Comparative analysis of wall clock speedup across different decoding method for 1-best rescoring over a top-16 BPD lattice. This figure presents the relative wall clock speedup, normalized against the baseline autoregressive decoding. Speedup values are derived theoretically, aggregating precalculated inference times for each method. Base BPD denotes the vanilla BPD inference.}
%\vspace{-15pt}
%\label{fig:wallclock}
%\end{figure*}

\vspace{-5pt}
\subsection{Empirical evaluation} 
\vspace{-5pt}

\textbf{Block efficiency}\, \autoref{tab:fst_results} and \autoref{fig:parallel_ngram_bpd} demonstrate the impact of lattice rescoring on block efficiency across various tasks. Autoregressive neural, \ngram LM, and $p$-best \ngram BPD rescoring all demonstrate improvements in block efficiency, although gains are task-dependent.
%\vspace{-5pt}
\begin{itemize}
    \item \textbf{High initial block efficiency (LAMBADA, CNN/Daily)}: Both rescoring methods show little to no improvement, suggesting that vanilla BPD already produces high quality drafts.
    \item \textbf{Low initial block efficiency (SQuAD V1, SAMSUM, XSUM, NewsRoom)}: Both neural and \ngram rescoring lead to block efficiency gains, particularly with neural LMs achieving the best performance in some cases. This suggests that rescoring helps refine predictions and navigate the lattice more efficiently in these scenarios.
\end{itemize}

%\vspace{-5pt}
%\autoref{fig:c4_indomain} shows the differences of rescoring methods when comparing domain-specific training to C4 training. 
 
%  \theertha{adrian: if we are combining the Batched BPD and ngram rescoring, should results move here?}
 
% \paragraph{Autoregressive Models vs \ngram} While \ngrams offer computational efficiency, neural models provide more nuanced improvements in specific tasks. For instance, the 61M neural model achieves the best performance on SAMSUM, showcasing its ability to capture complex sequential patterns beyond simple \ngrams. However, \ngrams show more consistent improvements across several tasks, making them a valuable alternative under limited computational resources.

% \begin{table}[t] \tiny \addtolength{\tabcolsep}{-3.2pt}
\begin{wraptable}{r}{0.54\textwidth}\tiny
\vspace{-12pt}
\centering
\caption{Wins, ties, and losses of 61M parameter neural-rescored drafts and vanilla drafts. ``\% Repair'' corresponds to instances where the rescored draft eliminates repetition and ``\% Regress'' corresponds to instances where the rescored draft introduces repetition.}
\label{tab:repetition_error_analysis}
\addtolength{\tabcolsep}{-4.5pt}
\begin{tabular}{c|c|c|c|c|c|c|c}
\toprule
\multirow{2.5}{*}{\textbf{Dataset}} & \multirow{2.5}{*}{\textbf{Ties}} & \multicolumn{3}{c|}{\textbf{Win}} & \multicolumn{3}{c}{\textbf{Loss}} \\ \cmidrule{3-8}
 &  & \textbf{Total} & \textbf{\% Repair} & \textbf{\% Regress} & \textbf{Total} & \textbf{\% Repair} & \textbf{\% Regress} \\ \midrule
LAMBADA & 631.5K & 5804 & 27.95 & 0.05 & 9466 & 2.01 & 0.06 \\
SQuAD V1 & 104.4K & 1624 & 12.68 & 8.13 & 6325 & 2.53 & 12.28 \\
CNN/Daily & 965.0K & 5928 & 23.20 & 0.67 & 17863 & 3.19 & 0.48 \\
SAMSum & 12.1K & 2462 & 17.91 & 23.56 & 867 & 18.57 & 16.72 \\
MultiNews & 1.45M & 294856 & 44.41 & 7.45 & 50209 & 22.21 & 5.37 \\
XSUM & 262.0K & 36010 & 29.87 & 0.77 & 6826 & 4.19 & 10.99 \\
NewsRoom & 251.3K & 79710 & 66.23 & 0.60 & 6492 & 2.85 & 7.39 \\  \bottomrule
\end{tabular}%
\vspace{-10pt}
% \end{table}
\end{wraptable}

\vspace{-5pt}
\textbf{Repairing repetitions}\, In \textbf{\Autoref{subsec:consec_repetition}}, we note that vanilla block drafts are prone to token-level repetition and that rescoring with a simple language model reduces the incidence of this. Although rescoring reduces repetition overall in drafts, is this driving improvements in block efficiency? To answer this, we compared the drafts generated by greedy rescoring with the 61M parameter neural rescorer against vanilla drafts. Time step instances were considered wins/ties/losses based on the accepted prefix length of the rescored draft vs. vanilla draft.  \autoref{tab:repetition_error_analysis} displays the win frequency across tasks along with the percentage of wins/losses attributed to introducing/eliminating repetition.

Note that in the tasks where rescoring improves block efficiency the most, NewsRoom and MultiNews, a high percentage of those repaired instances are driven by fixing erroneously repeated tokens. In fact, for MultiNews, 66.23\% of block drafts are improved through repetition repair. We also evaluated the performance of rescoring with in-domain trained rescoring LMs, but found that they tended to perform no better than C4-trained LMs (\textbf{\Autoref{sec:app_indomain_eval}}).

\vspace{-5pt}
\section{Conclusion} \label{sec:conclu}
\vspace{-5pt}
This paper presents a comprehensive analysis of BPD, highlighting its predictive dynamics and proposing methods to refine the generation of block drafts. Our study offers insights into BPD's behavior, particularly the tendency for drafts to contain consecutive repetitions and its heads to exhibit varying confidence levels in predictions. We introduce a novel measure, oracle top-$k$ block efficiency, to explore potential improvements in block efficiency. Two algorithms are proposed for generating higher quality drafts: one for local rescoring with small neural models (i.e., neural BPD) and another for global rescoring with an \ngram LM and generating multiple drafts (i.e. $p$-best \ngram BPD). These algorithms leverage the strengths of both blockwise parallel LMs and small rescoring models to reduce average decoding latency, pushing the boundaries of efficient text generation with BPD. We believe that this paper lays the groundwork for future exploration in optimizing LM decoding speed.

\clearpage
\bibliography{ref}

\begin{thebibliography}{10}

\bibitem{openfst}
Cyril Allauzen, Michael Riley, Johan Schalkwyk, Wojciech Skut, and Mehryar
  Mohri.
\newblock Openfst: A general and efficient weighted finite-state transducer
  library: (extended abstract of an invited talk).
\newblock In {\em Implementation and Application of Automata: 12th
  International Conference, CIAA 2007, Praque, Czech Republic, July 16-18,
  2007, Revised Selected Papers 12}, pages 11--23. Springer, 2007.

\bibitem{jax2018github}
James Bradbury, Roy Frostig, Peter Hawkins, Matthew~James Johnson, Chris Leary,
  Dougal Maclaurin, George Necula, Adam Paszke, Jake Vander{P}las, Skye
  Wanderman-{M}ilne, and Qiao Zhang.
\newblock {JAX}: composable transformations of {P}ython+{N}um{P}y programs,
  2018.

\bibitem{gpt3}
Tom Brown, Benjamin Mann, Nick Ryder, Melanie Subbiah, Jared~D Kaplan, Prafulla
  Dhariwal, Arvind Neelakantan, Pranav Shyam, Girish Sastry, Amanda Askell,
  et~al.
\newblock Language models are few-shot learners.
\newblock {\em Advances in neural information processing systems},
  33:1877--1901, 2020.

\bibitem{medusa}
Tianle Cai, Yuhong Li, Zhengyang Geng, Hongwu Peng, Jason~D Lee, Deming Chen,
  and Tri Dao.
\newblock Medusa: Simple llm inference acceleration framework with multiple
  decoding heads.
\newblock {\em arXiv preprint arXiv:2401.10774}, 2024.

\bibitem{speculative_decode2}
Charlie Chen, Sebastian Borgeaud, Geoffrey Irving, Jean-Baptiste Lespiau,
  Laurent Sifre, and John Jumper.
\newblock Accelerating large language model decoding with speculative sampling.
\newblock {\em arXiv preprint arXiv:2302.01318}, 2023.

\bibitem{chi2020align}
Ethan~A Chi, Julian Salazar, and Katrin Kirchhoff.
\newblock Align-refine: Non-autoregressive speech recognition via iterative
  realignment.
\newblock {\em arXiv preprint arXiv:2010.14233}, 2020.

\bibitem{palm}
Aakanksha Chowdhery, Sharan Narang, Jacob Devlin, Maarten Bosma, Gaurav Mishra,
  Adam Roberts, Paul Barham, Hyung~Won Chung, Charles Sutton, Sebastian
  Gehrmann, et~al.
\newblock Palm: Scaling language modeling with pathways.
\newblock {\em arXiv preprint arXiv:2204.02311}, 2022.

\bibitem{flashattention}
Tri Dao, Dan Fu, Stefano Ermon, Atri Rudra, and Christopher R{\'e}.
\newblock Flashattention: Fast and memory-efficient exact attention with
  io-awareness.
\newblock {\em Advances in Neural Information Processing Systems},
  35:16344--16359, 2022.

\bibitem{dettmers2022gpt3}
Tim Dettmers, Mike Lewis, Younes Belkada, and Luke Zettlemoyer.
\newblock Gpt3. int8 (): 8-bit matrix multiplication for transformers at scale.
\newblock {\em Advances in Neural Information Processing Systems},
  35:30318--30332, 2022.

\bibitem{elbayad2019depth}
Maha Elbayad, Jiatao Gu, Edouard Grave, and Michael Auli.
\newblock Depth-adaptive transformer.
\newblock {\em arXiv preprint arXiv:1910.10073}, 2019.

\bibitem{multi_news}
Alexander~R Fabbri, Irene Li, Tianwei She, Suyi Li, and Dragomir~R Radev.
\newblock Multi-news: A large-scale multi-document summarization dataset and
  abstractive hierarchical model.
\newblock {\em arXiv preprint arXiv:1906.01749}, 2019.

\bibitem{SAMSum}
Bogdan Gliwa, Iwona Mochol, Maciej Biesek, and Aleksander Wawer.
\newblock Samsum corpus: A human-annotated dialogue dataset for abstractive
  summarization.
\newblock {\em arXiv preprint arXiv:1911.12237}, 2019.

\bibitem{newsroom}
Max Grusky, Mor Naaman, and Yoav Artzi.
\newblock {N}ewsroom: A dataset of 1.3 million summaries with diverse
  extractive strategies.
\newblock In Marilyn Walker, Heng Ji, and Amanda Stent, editors, {\em
  Proceedings of the 2018 Conference of the North {A}merican Chapter of the
  Association for Computational Linguistics: Human Language Technologies,
  Volume 1 (Long Papers)}, pages 708--719, New Orleans, Louisiana, June 2018.
  Association for Computational Linguistics.

\bibitem{nonar}
Jiatao Gu, James Bradbury, Caiming Xiong, Victor~OK Li, and Richard Socher.
\newblock Non-autoregressive neural machine translation.
\newblock {\em arXiv preprint arXiv:1711.02281}, 2017.

\bibitem{cnn_sum}
Karl~Moritz Hermann, Tomas Kocisky, Edward Grefenstette, Lasse Espeholt, Will
  Kay, Mustafa Suleyman, and Phil Blunsom.
\newblock Teaching machines to read and comprehend.
\newblock {\em Advances in neural information processing systems}, 28, 2015.

\bibitem{chinchilla}
Jordan Hoffmann, Sebastian Borgeaud, Arthur Mensch, Elena Buchatskaya, Trevor
  Cai, Eliza Rutherford, Diego de~Las~Casas, Lisa~Anne Hendricks, Johannes
  Welbl, Aidan Clark, et~al.
\newblock An empirical analysis of compute-optimal large language model
  training.
\newblock {\em Advances in Neural Information Processing Systems},
  35:30016--30030, 2022.

\bibitem{nucleus_sampling}
Ari Holtzman, Jan Buys, Li~Du, Maxwell Forbes, and Yejin Choi.
\newblock The curious case of neural text degeneration.
\newblock {\em arXiv preprint arXiv:1904.09751}, 2019.

\bibitem{jouppi2017datacenter}
Norman~P Jouppi, Cliff Young, Nishant Patil, David Patterson, Gaurav Agrawal,
  Raminder Bajwa, Sarah Bates, Suresh Bhatia, Nan Boden, Al~Borchers, et~al.
\newblock In-datacenter performance analysis of a tensor processing unit.
\newblock In {\em Proceedings of the 44th annual international symposium on
  computer architecture}, pages 1--12, 2017.

\bibitem{katz1987estimation}
Slava Katz.
\newblock Estimation of probabilities from sparse data for the language model
  component of a speech recognizer.
\newblock {\em IEEE transactions on acoustics, speech, and signal processing},
  35(3):400--401, 1987.

\bibitem{biglittlespec}
Sehoon Kim, Karttikeya Mangalam, Suhong Moon, Jitendra Malik, Michael~W
  Mahoney, Amir Gholami, and Kurt Keutzer.
\newblock Speculative decoding with big little decoder.
\newblock In {\em Thirty-seventh Conference on Neural Information Processing
  Systems}, 2023.

\bibitem{instructive_decoding}
Taehyeon Kim, Joonkee Kim, Gihun Lee, and Se-Young Yun.
\newblock Instructive decoding: Instruction-tuned large language models are
  self-refiner from noisy instructions.
\newblock In {\em The Twelfth International Conference on Learning
  Representations}, 2024.

\bibitem{kitaev2020reformer}
Nikita Kitaev, {\L}ukasz Kaiser, and Anselm Levskaya.
\newblock Reformer: The efficient transformer.
\newblock {\em arXiv preprint arXiv:2001.04451}, 2020.

\bibitem{moses}
Philipp Koehn, Hieu Hoang, Alexandra Birch, Chris Callison-Burch, Marcello
  Federico, Nicola Bertoldi, Brooke Cowan, Wade Shen, Christine Moran, Richard
  Zens, et~al.
\newblock Moses: Open source toolkit for statistical machine translation.
\newblock In {\em Proceedings of the 45th annual meeting of the association for
  computational linguistics companion volume proceedings of the demo and poster
  sessions}, pages 177--180. Association for Computational Linguistics, 2007.

\bibitem{speculative_decode}
Yaniv Leviathan, Matan Kalman, and Yossi Matias.
\newblock Fast inference from transformers via speculative decoding.
\newblock In {\em International Conference on Machine Learning}, pages
  19274--19286. PMLR, 2023.

\bibitem{contrastive_decoding}
Xiang~Lisa Li, Ari Holtzman, Daniel Fried, Percy Liang, Jason Eisner, Tatsunori
  Hashimoto, Luke Zettlemoyer, and Mike Lewis.
\newblock Contrastive decoding: Open-ended text generation as optimization.
\newblock {\em arXiv preprint arXiv:2210.15097}, 2022.

\bibitem{ma2023llm}
Xinyin Ma, Gongfan Fang, and Xinchao Wang.
\newblock Llm-pruner: On the structural pruning of large language models.
\newblock {\em arXiv preprint arXiv:2305.11627}, 2023.

\bibitem{pass}
Giovanni Monea, Armand Joulin, and Edouard Grave.
\newblock Pass: Parallel speculative sampling.
\newblock {\em arXiv preprint arXiv:2311.13581}, 2023.

\bibitem{xsum}
Shashi Narayan, Shay~B Cohen, and Mirella Lapata.
\newblock Don't give me the details, just the summary! topic-aware
  convolutional neural networks for extreme summarization.
\newblock {\em arXiv preprint arXiv:1808.08745}, 2018.

\bibitem{lambada}
Denis Paperno, Germ{\'a}n Kruszewski, Angeliki Lazaridou, Quan~Ngoc Pham,
  Raffaella Bernardi, Sandro Pezzelle, Marco Baroni, Gemma Boleda, and Raquel
  Fern{\'a}ndez.
\newblock The lambada dataset: Word prediction requiring a broad discourse
  context.
\newblock {\em arXiv preprint arXiv:1606.06031}, 2016.

\bibitem{lms_are_unsupervised_multitask_learners}
Alec Radford, Jeffrey Wu, Rewon Child, David Luan, Dario Amodei, Ilya
  Sutskever, et~al.
\newblock Language models are unsupervised multitask learners.
\newblock {\em OpenAI blog}, 1(8):9, 2019.

\bibitem{gopher}
Jack~W Rae, Sebastian Borgeaud, Trevor Cai, Katie Millican, Jordan Hoffmann,
  Francis Song, John Aslanides, Sarah Henderson, Roman Ring, Susannah Young,
  et~al.
\newblock Scaling language models: Methods, analysis \& insights from training
  gopher.
\newblock {\em arXiv preprint arXiv:2112.11446}, 2021.

\bibitem{c4}
Colin Raffel, Noam Shazeer, Adam Roberts, Katherine Lee, Sharan Narang, Michael
  Matena, Yanqi Zhou, Wei Li, and Peter~J. Liu.
\newblock Exploring the limits of transfer learning with a unified text-to-text
  transformer.
\newblock {\em arXiv e-prints}, 2019.

\bibitem{t5}
Colin Raffel, Noam Shazeer, Adam Roberts, Katherine Lee, Sharan Narang, Michael
  Matena, Yanqi Zhou, Wei Li, and Peter~J Liu.
\newblock Exploring the limits of transfer learning with a unified text-to-text
  transformer.
\newblock {\em The Journal of Machine Learning Research}, 21(1):5485--5551,
  2020.

\bibitem{squad}
Pranav Rajpurkar, Jian Zhang, Konstantin Lopyrev, and Percy Liang.
\newblock Squad: 100,000+ questions for machine comprehension of text.
\newblock {\em arXiv preprint arXiv:1606.05250}, 2016.

\bibitem{calm}
Tal Schuster, Adam Fisch, Jai Gupta, Mostafa Dehghani, Dara Bahri, Vinh Tran,
  Yi~Tay, and Donald Metzler.
\newblock Confident adaptive language modeling.
\newblock {\em Advances in Neural Information Processing Systems},
  35:17456--17472, 2022.

\bibitem{jhuws2006iwslt}
Wade Shen, Richard Zens, Nicola Bertoldi, and Marcello Federico.
\newblock The jhu workshop 2006 iwslt system.
\newblock In {\em Proceedings of the Third International Workshop on Spoken
  Language Translation: Evaluation Campaign}, 2006.

\bibitem{spector2023accelerating}
Benjamin Spector and Chris Re.
\newblock Accelerating llm inference with staged speculative decoding.
\newblock {\em arXiv preprint arXiv:2308.04623}, 2023.

\bibitem{bpd}
Mitchell Stern, Noam Shazeer, and Jakob Uszkoreit.
\newblock Blockwise parallel decoding for deep autoregressive models.
\newblock {\em Advances in Neural Information Processing Systems}, 31, 2018.

\bibitem{stolcke2000entropy}
Andreas Stolcke.
\newblock Entropy-based pruning of backoff language models.
\newblock {\em arXiv preprint cs/0006025}, 2000.

\bibitem{sun2023simple}
Mingjie Sun, Zhuang Liu, Anna Bair, and J~Zico Kolter.
\newblock A simple and effective pruning approach for large language models.
\newblock {\em arXiv preprint arXiv:2306.11695}, 2023.

\bibitem{spectr}
Ziteng Sun, Ananda~Theertha Suresh, Jae~Hun Ro, Ahmad Beirami, Himanshu Jain,
  and Felix Yu.
\newblock Spectr: Fast speculative decoding via optimal transport.
\newblock {\em arXiv preprint arXiv:2310.15141}, 2023.

\bibitem{gemini}
Gemini Team, Rohan Anil, Sebastian Borgeaud, Yonghui Wu, Jean-Baptiste Alayrac,
  Jiahui Yu, Radu Soricut, Johan Schalkwyk, Andrew~M Dai, Anja Hauth, et~al.
\newblock Gemini: a family of highly capable multimodal models.
\newblock {\em arXiv preprint arXiv:2312.11805}, 2023.

\bibitem{emergent_abilities_llms}
Jason Wei, Yi~Tay, Rishi Bommasani, Colin Raffel, Barret Zoph, Sebastian
  Borgeaud, Dani Yogatama, Maarten Bosma, Denny Zhou, Donald Metzler, et~al.
\newblock Emergent abilities of large language models.
\newblock {\em arXiv preprint arXiv:2206.07682}, 2022.

\bibitem{xiao2023smoothquant}
Guangxuan Xiao, Ji~Lin, Mickael Seznec, Hao Wu, Julien Demouth, and Song Han.
\newblock Smoothquant: Accurate and efficient post-training quantization for
  large language models.
\newblock In {\em International Conference on Machine Learning}, pages
  38087--38099. PMLR, 2023.

\bibitem{yao2022zeroquant}
Zhewei Yao, Reza Yazdani~Aminabadi, Minjia Zhang, Xiaoxia Wu, Conglong Li, and
  Yuxiong He.
\newblock Zeroquant: Efficient and affordable post-training quantization for
  large-scale transformers.
\newblock {\em Advances in Neural Information Processing Systems},
  35:27168--27183, 2022.

\bibitem{zhu2021long}
Chen Zhu, Wei Ping, Chaowei Xiao, Mohammad Shoeybi, Tom Goldstein, Anima
  Anandkumar, and Bryan Catanzaro.
\newblock Long-short transformer: Efficient transformers for language and
  vision.
\newblock {\em Advances in neural information processing systems},
  34:17723--17736, 2021.

\end{thebibliography}
\bibliographystyle{plain}

\clearpage
\appendix
\section{Broader Impact} \label{sec:broad}

Our research in advancing blockwise parallel decoding (BPD) for language models (LMs) paves the way for substantial improvements in language processing. This section outlines the broader implications of our work, focusing on key areas such as efficiency, scalability, and systematic impact.

\begin{itemize}
    \item \textbf{Inference efficiency:} We achieve further acceleration in inference with minimal increases in CPU overhead and memory bandwidth utilization. In particular, we show that our methods significantly boost the inference efficiency compared to the standard autoregressive inference.  % (\Autoref{fig:overall_app}).
    
    \item \textbf{Scalability:} The introduction of selecting top $p$ sequences from top $k$ predictions demonstrates notable scalability, hinting at accelerated inference with growing model sizes. This scalability is further explored through diverse experiments involving various \ngram{} and neural LM architectures.
    
    \item \textbf{Systematic impact:} The advancement of BPD promises identical generation quality to traditional autoregressive methods, but with significantly improved efficiency in decoding times%. 
    %In parallel with speculative decoding research, our approach can reshape the landscape of LLM, offering faster, more accessible AI solutions across various industries. 
    , which can be helpful in latency-critical applications. 
\end{itemize}

% \begin{figure}[h]
%     \centering
%     \includegraphics[width=0.55\linewidth]{figure/app_overall_result.pdf}
%     \caption{Overall performance for a 1.5B blockwise parallel LM on 7 different datasets \cite{lambada, squad, cnn_sum, SAMSum, multi_news, xsum, newsroom}. Block efficiency gains are given absolute. \% change in KV Cache I/O and FLOPS per token are relative to standard autoregressive decoding to our BPD methods. Details are described in \textbf{\Autoref{sec:app_memory}}.} 
%     \label{fig:overall_app}
% \end{figure}

\section{Limitation \& Future Work} \label{sec:app_limit}

\subsection{Limitation}

\paragraph{Extending beyond greedy decoding}

In this work, we focus on refining block drafts for greedy decoding, the decoding scheme considered in the original BPD paper \cite{bpd}. Extending this methodology to non-greedy sampling strategies is an area for future work. Leviathan et al.\,\cite{speculative_decode} demonstrates the feasibility of applying speculative decoding strategies to accelerate (non-greedy) sampling. Their approach underscores the universality of the verification process across diverse model architectures, contingent on the availability of draft token-level predictions and probabilities — a requirement met by (rescored) block candidates.

\paragraph{Scaling up to larger models}  We show that lattice rescoring can improve the block efficiency for a 1.5 billion parameter blockwise parallel LM. It remains an open question as to whether one would observe similar improvements in block efficiency with larger LMs.

\paragraph{Memory overhead during parallel verification} A notable challenge in our framework is the memory overhead encountered during parallel verification. The simultaneous processing of multiple token predictions, particularly for larger models and draft batches when coupled with lattice rescoring, can introduce high demands on memory. That said, this can be mitigated by the application of tree attention during verification \cite{spector2023accelerating}.

\paragraph{Naive drafting heads and training recipes}

The implementation of drafting heads within our BPD framework, while hewing close to the original BPD proposal, leaves room for architecture experimentation. The choice of architecture for the block drafter is crucial, and will likely affect the gains one observes from rescoring the block draft top-$k$ lattice. Future iterations could benefit from more sophisticated training methods, enhancing these heads' ability to navigate complex linguistic contexts and improving overall predictive performance.

\subsection{Future Work}

The evolution of the BPD framework will be pivotal in addressing its current constraints and broadening its utility. Key areas for future research include:

\begin{itemize}
    \item The intrinsic compatibility of our lattice rescoring method with alternative sampling strategies presents fertile ground for future work. This delineation not only enriches the discourse within efficient language model inference but also sets a trajectory for subsequent empirical endeavors.
    
    \item Scaling the blockwise parallel LM for compatibility with larger-scale LLMs, ensuring it remains effective and efficient as language models evolve.
    
    \item Advancing training methodologies for drafting heads, to bolster their predictive accuracy and contribution to the BPD process, thereby optimizing the framework for more complex language modeling tasks.

    \item Using the sequential entropy ordering of heads (\Autoref{fig:entropy_block}) as a possible halting condition during block draft head training, or to inform how a rescoring LM should be interpolated with the block lattice weights.
\end{itemize}

\section{Experiment Details}
\label{sec:app_architecture}

\subsection{Training objective for blockwise parallel LMs}

We minimized the following loss function to train blockwise parallel LMs:

\[
\mathcal{L}_{BPD} = \sum_{h=1}^{H} \lambda_h \mathcal{L}_{h}
\]

where \(H\) is the number of heads, \(\lambda_h\) is a non-negative scalar that weights the loss from head \(h\), and \(\mathcal{L}_{h}\) denotes the loss for each individual head:

\[
\mathcal{L}_{h} = -\sum_{x_{1 \ldots i}, y_{i+h}} \log p(y_{i+h} | x_{1 \ldots i})
\]

where \(x_{1 \ldots i}\) is the token sequence up to position \(i\), \(y_{i+h}\) is the ground truth token at position \(i+h\), and \(p(y_{i+h} | x_{1 \ldots i})\) is the probability of observing token \(y_{i+h}\) given the sequence \(x_{1 \ldots i}\) under the blockwise parallel LM. We trained all models in this work with \(\lambda_h = 1\). We leave tuning these hyperparameters, improving the block efficiency and quality of the blockwise parallel LM, as future work.

\subsection{Neural Model Details}

\begin{table}[ht]
\centering
\caption{Architecture hyperparameters for each of the transformer-based neural language models.}
\begin{tabular}{c|r|c|c|c}
\toprule
 Type & Model &  \# Layers & Embedding Dim & Hidden Dim \\ \midrule
 Blockwise Parallel Decoder & 1.5B & 18 & 1,536 & 12,288 \\ \midrule
 \multirow{3}{*}{Autoregressive Decoder} & 32M & 2 & 384 & 1,536 \\
 & 61M & 12 & 384 & 1,536 \\
 & 94M & 6 & 768 & 3,072 \\ \bottomrule
\end{tabular}
\label{tab:neural_architecture}
\end{table}

Each neural rescoring LM is a decoder-only transformer with learned absolute positional embeddings and twelve self-attention heads at each layer. The key architecture hyperparameters are given in \Autoref{tab:neural_architecture}. Aside from scale, the only difference between the blockwise parallel LM and neural rescoring models is the addition of the feedforward neural networks and eight additional block prediction heads. Note that the number of parameters for each of these models also includes the embedding table.

Each model was pretrained on the English C4 corpus for 200K iterations with a batch size of $2^{20} \approx 1M$ tokens per batch. Dropout was not applied. For the blockwise parallel LM, all heads were trained jointly. The pretraining for the blockwise parallel LMs took about 47 hours on 128 TPUv3 units.

For downstream tasks, models were finetuned for a maximum 100K iterations with a batch size of two examples with maximum sequence length of 2048. Maximum learning rate was fixed to $10^{-4}$ for all runs, with a cosine learning rate schedule. Checkpoints were selected based on heldout set model performance. Interpolation weight for all rescoring models was tuned for block efficiency on 100 randomly selected examples from the evaluation set for each task, and performance was reported on the remainder of the evaluation set.

\subsection{\ngram Details}
\label{sec:app_ngram_details}

All \ngram{} LMs in this work are Katz backoff \ngram LMs \cite{katz1987estimation} fit on the train split of the GPT3 subword-tokenized English C4 corpus with \ngram order $\in \{2, 4\}$. We apply entropy pruning \citep{stolcke2000entropy} to reduce model size to a maximum of 100 million \ngrams per model, and ensure that each trigram is observed at least twice and each 4-gram is observed at least four times. Preprocessing of the text is identical to that used to train neural LMs.

\subsection{Datasets}

\begin{itemize}
    \item \textbf{LAMBADA (LAnguage Modeling Broadened to Account for Discourse Aspects)}: A collection of narrative passages designed to test the understanding of long-range dependencies in language models, where the task involves predicting the last word of a passage based on the full context \cite{lambada}.
    
    \item \textbf{SQuAD V1 (Stanford Question Answering Dataset)}: A reading comprehension dataset that features questions based on Wikipedia articles, with answers located within the text \cite{squad}.
    
    \item \textbf{CNN/DailyMail}: This dataset includes news articles paired with human-written summaries, mainly used to evaluate the summarization capabilities of language models, particularly in abstractive summarization \cite{cnn_sum}.
    
    \item \textbf{SAMSum (Semi-Automatic Machine Summarization)}: Focuses on abstractive summarization using news articles and machine-generated summaries, testing models' abilities to refine and improve existing summaries \cite{SAMSum}.
    
    \item \textbf{MultiNews}: Comprises news articles from diverse sources for abstractive summarization tasks, evaluating models on handling different writing styles and topics \cite{multi_news}.
    
    \item \textbf{XSUM}: Contains scientific documents and summaries, challenging language models to process complex scientific information and language \cite{xsum}.

    \item \textbf{NewsRoom}: A dataset of news articles aimed at assessing the factual accuracy and information extraction capabilities of models in generating summaries \cite{newsroom}.
\end{itemize}

All datasets were tokenized using the 50,257 GPT3 subword vocabulary \cite{gpt3}.

\paragraph{Templates}

We used the following prompts during model finetuning and inference.

\begin{itemize}
  \item \textbf{SQuAD}: "question: [question] context: [context]"
  \item \textbf{CNN/DailyMail}: "summarize: [text]"
  \item \textbf{SAMSum}: "Here is a dialogue: [text]$\backslash$nWrite a short summary!"
  \item \textbf{MultiNews}: "Write a summary based on this article: [text]"
  \item \textbf{XSUM}: "Summarize: [text]"
  \item \textbf{NewsRoom}: "Please write a short summary for the following article: [title] [text]"
\end{itemize}

% \newpage
\section{Rescoring with in-domain language models}
\label{sec:app_indomain_eval}

\begin{table}[ht]\small
\centering
\caption{Block efficiency from rescoring with in-domain trained rescoring models for 2-gram and 61M parameter neural rescorer.}
\label{tab:indomain_block_efficiency}
\begin{tabular}{l|c|c|c|c}
\toprule
\multirow{2}{*}{Dataset} & \multicolumn{2}{c|}{2-gram}& \multicolumn{2}{c}{neural-61M} \\\cmidrule{2-5}
& C4 & In-domain & C4 & In-domain \\ \midrule
SQuAD V1 & 2.09 & 2.04 & 2.10 & 2.06 \\
CNN/Daily & 1.73 & 1.73 & 1.73  & 1.72 \\
SAMSUM & 1.31 & 1.22 & 1.39 & 1.24 \\
MultiNews & 1.13 & 1.14 & 1.25 & 1.16 \\
XSUM & 1.17 & 1.18 & 1.23 & 1.14 \\
NewsRoom & 1.20 & 1.22 & 1.29 & 1.11 \\
\bottomrule
\end{tabular}%
\end{table}

We found that in-domain rescorers performed no better than rescorers only trained on C4. We suspect this is due to a lack of sufficient finetuning data and that the main benefit of rescoring comes from discouraging unnatural artifacts such as repetition from the original BPD draft. \Autoref{tab:indomain_block_efficiency} shows block efficiency after rescoring using in-domain models for all tasks besides language modeling.

Neural rescorers were finetuned from C4-pretrained checkpoints. \ngram{} models were trained from scratch, and unseen vocabulary was added as unigram arcs with trivial weight (negative log probability of 1000.0). This was done to ensure that all paths through the lattice were assigned non-zero probability by the \ngram{} model. In previous experiments, we also tried interpolating the in-domain \ngram{} model with a unigram model trained on C4, and observed similar performance as simply adding unseen unigrams.

%%%%%%%%%%%%%%%%%%%%%%%%%%%%%%%%%%%%%%%%%%%%%%%%%%%%%%%%%%%%%%%%%%%%%%%%%%%%%%%
%%%%%%%%%%%%%%%%%%%%%%%%%%%%%%%%%%%%%%%%%%%%%%%%%%%%%%%%%%%%%%%%%%%%%%%%%%%%%%%

\section{Interpolation weights tuned per task}

We tuned the interpolation weight, $\alpha$ for the 94M parameter neural and 4-gram LM rescorers, and then used this weight to rescore with all other models of that same class. 100 examples from each task's heldout set were set aside for tuning, to maximize block efficiency. The remainder of examples were used for evaluation. We swept over $\alpha \in \{0.1, 0.5, 0.75, 0.9, 1.0, 1.1, 1.5, 2.0, 5.0, 10.0\}$.

Note that for tasks where lattice rescoring was unhelpful, the interpolation weight, $\alpha$ is tuned to place much higher weight on the block draft logits (\Autoref{tab:tuned_interpolation_weight}). This is a signal that the rescorer does not provide additional information over the original block draft heads.

\begin{table}[h]\small
\centering
\caption{Tuned interpolation weight per task for neural and \ngram rescoring.}
\label{tab:tuned_interpolation_weight}
\begin{tabular}{l|c|c}
\toprule
\textbf{Dataset} & \textbf{Neural} & \textbf{\ngram} \\\midrule
LAMBADA & 0.1 & 0.1 \\
SQuAD V1 & 1.0 & 0.75 \\
SAMSum & 5.0 & 1.5 \\
CNN/Daily & 0.1 & 0.1 \\
MultiNews & 5.0 & 2.0 \\
XSUM & 1.5 & 1.1 \\
NewsRoom & 5.0 & 2.0 \\
\bottomrule
\end{tabular}%
\end{table}

% \newpage
\section{Local rescoring impact on block efficiency}

\begin{table*}[t]    \centering \small % \addtolength{\tabcolsep}{-0.5pt}
    \caption{Block efficiency after rescoring of the block lattice. Green circles (\textcolor{green!60!black}{$\CIRCLE$}) indicate improvement over the Baseline (BPD), with the percentage changes in block efficiency shown in brackets relative to the Baseline. Red circles (\textcolor{red!60!black}{$\CIRCLE$}) denote no improvement.}
    \resizebox{\textwidth}{!}{%
    \begin{tabular}{c|c|c|c|c|c|c|c|c|c|c} \toprule
        %         \multirow{2}{*}{Task} & \multirow{2}{*}{Dataset}   & Baseline & Local (new) & \multicolumn{3}{c|}{Global (new)} &  \multirow{2}{*}{Oracle (k=16)}   \\
        %   &   & BPD & neural-61M BPD & 4-gram BPD &  16-best 0-gram BPD & 16-best 4-gram BPD &  \\
    
        \multirow{2}{*}{Task} & \multirow{2}{*}{Dataset} & Baseline & \multicolumn{3}{c|}{Global rescoring} & \multicolumn{3}{c|}{Local rescoring} & \multirow{2}{*}{Oracle (k=2)} & \multirow{2}{*}{Oracle (k=16)} \\
        & & BPD & 2-gram BPD & 3-gram BPD & 4-gram BPD & neural-32M BPD & neural-61M BPD & neural-94M BPD &  & \\ \midrule
        LM & LAMBADA   & \underline{\textbf{3.12}} & 3.06 (-1.92\%) \textcolor{red!60!black}{$\CIRCLE$} & 3.05 (-2.24\%) \textcolor{red!60!black}{$\CIRCLE$} & 3.05 (-2.24\%) \textcolor{red!60!black}{$\CIRCLE$} & 3.08 (-1.28\%) \textcolor{red!60!black}{$\CIRCLE$} & 3.10 (-0.64\%) \textcolor{red!60!black}{$\CIRCLE$} & 3.05 (-2.24\%) \textcolor{red!60!black}{$\CIRCLE$} & 3.22 & 3.67 \\ \midrule
        QA & SQuAD V1  & 2.08 & 2.09 (+0.48\%) \textcolor{green!60!black}{$\CIRCLE$} & 2.08 (0.00\%)  \textcolor{red!60!black}{$\CIRCLE$} & 2.07 (-0.48\%) \textcolor{red!60!black}{$\CIRCLE$} & \underline{\textbf{2.10 (+0.96\%)}} \textcolor{green!60!black}{$\CIRCLE$} & \underline{\textbf{2.10 (+0.96\%)}} \textcolor{green!60!black}{$\CIRCLE$} & 2.07 (-0.48\%) \textcolor{red!60!black}{$\CIRCLE$} & 2.16 & 2.45\\ \midrule
        \multirow{2}{*}{S-SUM} & CNN/Daily & \underline{\textbf{1.74}} & 1.73 (-0.57\%) \textcolor{red!60!black}{$\CIRCLE$} & 1.73 (-0.57\%) \textcolor{red!60!black}{$\CIRCLE$} & 1.73 (-0.57\%) \textcolor{red!60!black}{$\CIRCLE$} & 1.73 (-0.57\%) \textcolor{red!60!black}{$\CIRCLE$} & 1.73 (-0.57\%) \textcolor{red!60!black}{$\CIRCLE$} & 1.73 (-0.57\%) \textcolor{red!60!black}{$\CIRCLE$} & 1.84 & 2.26 \\ 
        & SAMSum    & 1.27 & 1.31 (+3.15\%) \textcolor{green!60!black}{$\CIRCLE$} & 1.31 (+3.15\%) \textcolor{green!60!black}{$\CIRCLE$} & 1.29 (+1.57\%) \textcolor{green!60!black}{$\CIRCLE$} & 1.33 (+4.72\%) \textcolor{green!60!black}{$\CIRCLE$} & \underline{\textbf{1.39 (+9.45\%)}} \textcolor{green!60!black}{$\CIRCLE$} & 1.21 (-4.72\%) \textcolor{red!60!black}{$\CIRCLE$} & 1.23  & 1.95 \\ \midrule
        \multirow{3}{*}{L-SUM} & MultiNews & 1.10 & 1.13 (+2.73\%) \textcolor{green!60!black}{$\CIRCLE$} & 1.13 (+2.73\%) \textcolor{green!60!black}{$\CIRCLE$} & 1.12 (+1.82\%) \textcolor{green!60!black}{$\CIRCLE$} & \underline{\textbf{1.25 (+13.64\%)}} \textcolor{green!60!black}{$\CIRCLE$} & \underline{\textbf{1.25 (+13.64\%)}} \textcolor{green!60!black}{$\CIRCLE$} & 1.20 (+9.09\%) \textcolor{green!60!black}{$\CIRCLE$} & 1.13  & 1.43 \\
        & XSUM      & 1.13 & 1.17 (+3.54\%) \textcolor{green!60!black}{$\CIRCLE$} & 1.17 (+3.54\%) \textcolor{green!60!black}{$\CIRCLE$} & 1.16 (+2.65\%) \textcolor{green!60!black}{$\CIRCLE$} & 1.18 (+4.42\%) \textcolor{green!60!black}{$\CIRCLE$} & \underline{\textbf{1.23 (+8.85\%)}} \textcolor{green!60!black}{$\CIRCLE$} & 1.17 (+3.54\%) \textcolor{green!60!black}{$\CIRCLE$} & 1.17 & 1.55\\
        & NewsRoom  & 1.08 & 1.20 (+11.11\%) \textcolor{green!60!black}{$\CIRCLE$} & 1.18 (+9.26\%) \textcolor{green!60!black}{$\CIRCLE$} & 1.18 (+9.26\%) \textcolor{green!60!black}{$\CIRCLE$} & \underline{\textbf{1.29 (+19.44\%)}} \textcolor{green!60!black}{$\CIRCLE$} & \underline{\textbf{1.29 (+19.44\%)}} \textcolor{green!60!black}{$\CIRCLE$} & 1.17 (+8.33\%) \textcolor{green!60!black}{$\CIRCLE$} & 1.15 & 1.50 \\
    \bottomrule
    \end{tabular}
    }
    \vspace{-10pt}
    \label{tab:rescoring_results}
\end{table*}

\autoref{tab:rescoring_results} reveals the impact of different rescoring methods on the block efficiency of the block lattice, offering valuable insights into their effectiveness across diverse tasks and models, supporting the investigations in \textbf{\Autoref{sec:algorithms}}.

\begin{itemize}
    \item \textbf{Limited improvement for high baselines}: For tasks with already high initial block efficiency (LAMBADA, CNN/DailyMail), rescoring offers minimal or even negative changes in block efficiency compared to the baseline BPD system. This suggests that for tasks where standard BPD already achieves significant speed improvements, there is limited room for further gains through rescoring.
    
    \item \textbf{Efficacy for poor baselines}: In tasks with lower initial block efficiency (SQuAD V1, XSUM, NewsRoom), rescoring using both \ngram and neural language models results in increased block efficiency. Notably, neural rescoring with larger models (61M and 94M parameters) achieves the highest efficiency gains in these tasks, reaching up to 19.44\% improvement in NewsRoom. These results highlight the potential of rescoring to refine predictions and enhance efficiency for models exhibiting calibration issues.
    
    \item \textbf{Task-specific effectiveness}: The level of improvement from rescoring varies across different summarization tasks (MultiNews, XSUM, NewsRoom). While all show positive responses, NewsRoom exhibits the largest gains, suggesting that the effectiveness of rescoring is task-dependent.
    
    \item \textbf{Comparison with oracle efficiency}: The `Oracle' columns present the upper bound achievable if only the most likely token at each step is chosen with perfect hindsight (k=2 and k=16). While significant gaps remain between current results and the oracle, the observed improvements from rescoring demonstrate progress towards closing this efficiency gap.
\end{itemize}

Overall, these findings suggest that local rescoring methods can be a valuable tool for enhancing BPD efficiency, particularly for models with less calibrated predictions. Further exploration of advanced rescoring strategies, especially in conjunction with larger neural language models, holds promise for achieving even closer-to-oracle efficiency levels.

% \newpage

% \section{Local Rescoring via Neural Models vs Parallel BPD via Global \ngram Rescoring}

% \adrian{Drop this section? This is basically discussed in the last paragraph of section 2.2.}
% % \theertha{taehyeon: I think it would be good to update the language here to reflect Section 7 (batch to parallel etc) or remove it altogether. }
% This section supports the detailed explanations on two algorithms in \textbf{\Autoref{sec:algorithms}}: local rescoring via neural models and parallel BPD via global n-gram rescoring.

% \begin{itemize}
%     \item \textbf{Local rescoring via neural models} typically refers to a process where predictions are adjusted at each individual step of the sequence generation. A local rescoring algorithm  takes the initial predictions from the model and refine them using additional information or a secondary model.
    
%     \item \textbf{Parallel BPD via global \ngram rescoring}, on the other hand, implies a more comprehensive approach where the entire sequence or multiple sequences are considered together. The `global' aspect means that the rescoring process takes into account the whole sequence's context to find the optimal path, rather than making decisions based on individual tokens. Parallel batched processing indicates that multiple sequences are being evaluated and rescored in one go, which is often more efficient. \ngram rescoring uses the statistical properties of \ngrams to predict the probability of a sequence, allowing for the consideration of the sequence as a whole rather than as isolated predictions.
% \end{itemize}

\section{Ablation on the number of heads in the blockwise parallel LM}

\autoref{tab:head} summarizes the block efficiency for different head configurations across various language tasks with the same settings discussed in \autoref{tab:baseline}.

\begin{itemize}
    \item \textbf{General trend}: Both performance and block efficiency tend to increase with the number of heads, up to a point. This suggests that using more heads allows the model to capture richer contextual information and make more accurate predictions.
    
    \item \textbf{Efficiency trade-off}: While increasing heads generally improves block efficiency, it also increases the memory for verification stages. Therefore, the optimal number of heads depends on the balance between desired block efficiency and available resources.
\end{itemize}

\begin{table}[h] \small
\centering
\caption{Test performance per task. Test performance of each finetuned model and block efficiency are shown as a function of heads ($h \in {3,6,9}$). Tasks inclue Language Modeling (LM), extractive Question Answering (QA), and both Long and Short Summarization (L-Sum \& S-Sum). The metric for LM is perplexity, for QA is exact match, and for all the remaining (summarization) tasks, the metric is ROUGE-L.}
\label{tab:head}
\begin{tabular}{@{}c|l|c|ccc@{}}
\toprule
\multirow{2}{*}{Task} & \multirow{2}{*}{Dataset} & \multirow{2}{*}{Performance} & \multicolumn{3}{c}{\# of Heads ($h$)} \\ 
                              &           & & 3 & 6 & 9 \\ \midrule
LM                   & LAMBADA       & 7.88 & 1.79 & 2.84 & 3.12 \\ \midrule
QA                   & SQuAD V1       & 57.60 & 1.53 & 2.03 & 2.08 \\ \midrule
\multirow{2}{*}{S-SUM} & CNN/Daily     & 39.85 & 1.60 & 1.71 & 1.74 \\
                     & SAMSUM        & 37.66 & 1.18 & 1.25 & 1.27 \\ \midrule
\multirow{3}{*}{L-SUM} & MultiNews     & 23.08 & 1.08 & 1.08 & 1.10 \\
                     & XSUM          & 52.15 & 1.11 & 1.12 & 1.13 \\ 
                     & NewsRoom      & 39.85 & 1.07 & 1.08 & 1.08 \\ 
\bottomrule
\end{tabular}
% \vspace{-15pt}
\end{table}

\section{Practical efficiency of rescoring block drafts} \label{sec:app_memory}

To enhance our understanding of block rescoring within the realm of contemporary deep learning hardware environments, we present an in-depth examination focused on TPU/GPU utilization and the overhead incurred by \ngram{} rescoring. This analysis is divided into two parts: (1) an analysis of block rescoring through the lens of TPU/GPU utilization, and (2) empirical benchmarks of \ngram{} lattice rescoring. The major takeaways are as follows.

\paragraph{Memory bandwidth (HBM $\Leftrightarrow$ SRAM)}
A critical factor in the performance of deep learning applications is the efficient management of memory bandwidth between High Bandwidth Memory (HBM) and Static Random Access Memory (SRAM) \cite{flashattention}. Increasing the block efficiency via the block lattice rescoring reduces the average per token parameter and key-value cache I/O that needs to be communicated from HBM to SRAM.

\paragraph{Overhead in \ngram{} rescoring}
\ngram{} rescoring is actually quite efficient. For the size of lattices we consider in this work, moving the lattice from HBM to DRAM, performing n-best \ngram{} rescoring, and moving the n-best paths back to HBM requires no more than 2 ms per lattice.

\subsection{Hardware utilization}

We compare our approach against traditional Autoregressive LMs across several metrics (\Autoref{tab:tpu_gpu}):

\begin{table}[ht]
\caption{Comparative analysis of per decoded token efficiency metrics across block rescoring methods and the standard autoregressive LM (batch size=1). This table shows the average block efficiency, parameter I/O, key-value (KV) cache I/O at varying sequence lengths, and FLOPS—evaluated on a per-token basis with batch size 1.}
\centering
\resizebox{\textwidth}{!}{%
\begin{tabular}{l|cccccc}
\toprule
Component & Autoregressive & Base BPD & 4-gram BPD & Neural-61M BPD & 16-best 0-gram BPD & 16-best 4-gram BPD \\
\midrule
Avg. Block Efficiency & 1.000 & 1.646 & 1.657 & 1.724 & 1.717 & 1.797 \\
Parameter I/O (GB) & 3.000 & 1.823 & 1.811 & 1.811 & 1.747 & 1.669 \\ 
KV Cache I/O (GB) - Seq\_len 128 & 0.113 & 0.074 & 0.073 & 0.076 & 0.140 & 0.134 \\ 
KV Cache I/O (GB) - Seq\_len 512 & 0.453 & 0.280 & 0.278 & 0.290 & 0.338 & 0.323 \\ 
KV Cache I/O (GB) - Seq\_len 1024 & 0.906 & 0.555 & 0.552 & 0.574 & 0.602 & 0.575 \\ 
KV Cache I/O (GB) - Seq\_len 2048 & 1.812 & 1.106 & 1.098 & 1.144 & 1.129 & 1.079 \\ 
FLOPS (T) & 0.931 & 0.57 & 0.567 & 0.635 & 0.621 & 0.593 \\
\bottomrule
\end{tabular}
} \label{tab:tpu_gpu}
\end{table}

\paragraph{Memory bandwidth and compute efficiency}
The block rescoring variants achieve significant reductions in Parameter I/O and KV Cache I/O compared to autoregressive decoding, suggesting BPD methods' ability to reducing inference times by mitigating the primary latency bottleneck—memory bandwidth. Advances in TPU/GPU architecture ensure that an increase in FLOPS per token has a minimal effect on latency, confirming our strategy's capacity to navigate the complexities of memory bandwidth efficiently.

\paragraph{Comparative latency impact}
A consistent decrease in memory bandwidth utilization across blockwise parallel LMs, including those leveraging LM rescoring and parallel processing strategies, illustrates our approach's contribution to accelerating inference speed. This underscores the practicality and applicability of our enhancements in promoting more efficient language model inference within state-of-the-art computational frameworks.

\subsection{Overhead of n-gram rescoring}
While the majority of computational efforts in block rescoring are dedicated to TPU/GPU utilization, the implementation of n-gram rescoring introduces additional overheads. These are primarily attributed to CPU computations and the data transfer between the CPU and HBM. This section provides a comprehensive examination of these overheads, drawing on benchmarks from rescoring experiments with a 4-gram C4 LM.

\paragraph{Benchmarks for 4-gram C4 LM rescoring} We conducted benchmarks on rescoring lattices with a 4-gram C4 LM of  $\approx$100M n-grams. The average latency observed across 10 runs for different numbers of the shortest paths is summarized in \Autoref{tab:short_path}.

\begin{table}[ht]
\caption{Average latency for N-best rescoring an 8-time step lattice with 16 arcs per time step. N, the number of shortest paths, is varied from 1 to 16.} \label{tab:short_path}
\centering
\begin{tabular}{c|c}
\toprule
\# Shortest Paths & N-best Rescoring Latency (ms) \\
\midrule
1  & 1.630 \\ % 1.629543304 \\
2  & 1.751 \\ %1.750683784 \\
4  & 1.878 \\ % 1.878190041 \\
8  & 1.871 \\ % 1.870632172 \\
16 & 1.983 \\ % 1.982903481 \\
\bottomrule
\end{tabular}
\end{table}

Notably, rescoring with a large 4-gram LM averages less than 2 milliseconds for extracting up to 16 globally-best paths, despite the lattice containing approximately 4.29 billion possible paths. In our initial experiments, increasing the size of the \ngram{} LM had little effect on n-best rescoring latency, indicating that improvements to rescoring LM quality will incur little additional latency, provided that the rescoring LM fits within DRAM.

Latency is predominantly influenced by lattice size, particularly the number of top-k tokens per time step and the number of time steps, as depicted in \Autoref{tab:1_best}.

\begin{table}[ht]
\caption{1-best rescoring latency by the 4-gram C4 LM for varying lattice sizes.} \label{tab:1_best}
\centering
\begin{tabular}{c|c|c}
\toprule
Number of time steps & Top-k per time step & 1-best rescoring latency (ms) \\ 
\midrule
4 & 2 & 1.038   \\ % 1.03764534   \\ 
4 & 4 & 1.050  \\ % 1.050209999  \\ 
4 & 8 & 1.130  \\ % 1.130461693  \\ 
4 & 16& 1.237  \\ \midrule % 1.236653328  \\ \midrule
8 & 2 & 1.061  \\ % 1.061272621  \\
8 & 4 & 1.144  \\ % 1.143503189  \\
8 & 8 & 1.234  \\ % 1.23398304  \\
8 & 16 & 1.630 \\ \midrule % 1.629543304 \\ \midrule
16 & 2 & 1.102 \\ % 1.102280617 \\
16 & 4 & 1.206 \\ % 1.206278801 \\
16 & 8 & 1.558 \\ % 1.558089256 \\
16 & 16 & 2.174 \\ % 2.174425125 \\
\bottomrule
\end{tabular}
\end{table}

The benchmarks highlight the fact that the additional overhead introduced by \ngram{} rescoring, though present, should not significantly impact overall latency.

% \clearpage
% \input{checklist}

\end{document}